\pdfoutput=1
\PassOptionsToPackage{table}{xcolor}
\documentclass[11pt]{article}
\usepackage[final]{acl}

\usepackage[utf8]{inputenc}
\usepackage[T1]{fontenc}
\usepackage{CJKutf8}

\usepackage{amsmath}
\usepackage{amssymb}
\usepackage{mathtools}
\newcommand{\norm}[1]{\left\lVert #1 \right\rVert}
\usepackage{multirow}
\usepackage{pifont}

\usepackage{graphicx}
\usepackage{subcaption}
\usepackage{placeins}
\usepackage{booktabs}
\usepackage{float}
\usepackage{dblfloatfix}
\usepackage{caption}
\usepackage[skip=6pt]{caption}
\usepackage{tabularx}
\usepackage{ragged2e}

\usepackage{times}
\usepackage{latexsym}
\usepackage{url}
\usepackage{microtype}
\usepackage{inconsolata}
\usepackage{relsize}
\usepackage{array}
\usepackage{pifont}
\usepackage{longtable}
\usepackage{tipa}
\usepackage{algorithm}
\usepackage{algpseudocode}

\usepackage{tikz}
\usepackage{tikzscale}
\usetikzlibrary{bayesnet}
\definecolor{bratgreen}{HTML}{8ACE00}

\title{\textsc{PhoniTale}: Phonologically Grounded Mnemonic Generation \\for Typologically Distant Language Pairs}

\author{
\textbf{Sana Kang\textsuperscript{1}\thanks{These authors contributed equally to this work.}} \quad
\textbf{Myeongseok Gwon\textsuperscript{1}\footnotemark[1]} \quad
\textbf{Su Young Kwon\textsuperscript{1}\footnotemark[1]} \quad \\
\textbf{Jaewook Lee\textsuperscript{2}} \quad
\textbf{Andrew Lan\textsuperscript{2}} \quad
\textbf{Bhiksha Raj\textsuperscript{3}} \quad
\textbf{Rita Singh\textsuperscript{3}} \\
\textsuperscript{1}KAIST \quad
\textsuperscript{2}University of Massachusetts Amherst \quad
\textsuperscript{3}Carnegie Mellon University \\
  \texttt{%
    \begin{tabular}{>{\centering\arraybackslash}m{0.95\linewidth}}
      \{sanakang0615, myeongseok, suyoungkwon\}@kaist.ac.kr, \\
      \{jaewooklee, andrewlan\}@cs.umass.edu,
      \{bhiksha, rsingh\}@cs.cmu.edu
    \end{tabular}
  }
}

\begin{document}
\begin{CJK}{UTF8}{mj} 
\maketitle

\begin{abstract}
Vocabulary acquisition poses a significant challenge for second-language (L2) learners, especially when learning typologically distant languages such as English and Korean, where phonological and structural mismatches complicate vocabulary learning. Recently, large language models (LLMs) have been used to generate keyword mnemonics by leveraging similar keywords from a learner’s first language (L1) to aid in acquiring L2 vocabulary. However, most methods still rely on direct IPA-based phonetic matching or employ LLMs without phonological guidance. In this paper, we present \textsc{PhoniTale}, a novel cross-lingual mnemonic generation system that performs IPA-based phonological adaptation and syllable-aware alignment to retrieve L1 keyword sequence and uses LLMs to generate verbal cues. We evaluate \textsc{PhoniTale} through automated metrics and a short-term recall test with human participants, comparing its output to human-written and prior automated mnemonics. Our findings show that \textsc{PhoniTale} consistently outperforms previous automated approaches and achieves quality comparable to human-written mnemonics.

\end{abstract}

\section{Introduction}
\label{sec:introduction}

Vocabulary acquisition remains one of the most persistent challenges for second-language (L2) learners.
A classic—and surprisingly durable—strategy is \emph{keyword mnemonic}: learners associate a new L2 lexical item with a familiar first-language (L1) word or phrase whose pronunciation is similar, and then build a vivid verbal or visual scene that links the two \cite{atkinson1975keyword}. For example, a German learner might associate the word \emph{Flasche} (bottle) with the phonetically similar English word \emph{flashy}, forming the mnemonic \emph{a flashy bottle that stands out from the rest}. This technique leverages phonological similarity while establishing a memorable semantic connection between the L2 target and L1 knowledge~\cite{lee2023smartphone}.

Typically, the L1 phrase corresponding to a given L2 term to be memorized is manually designed; however, this is a laborious process that scales poorly, necessitating the development of automated mechanisms to compose these phrases. Methods for automated generation of such keywords began with \textsc{TransPhoner}, which leverages the International Phonetic Alphabet (IPA) \cite{IPA1949} and hand-crafted heuristics to retrieve pronunciation-similar L1 keyword for target English words, resulting in significant recall gains~\cite{savva2014transphoner}.

Leveraging these methods, recent studies employ large language models (LLMs) for automated keyword generation. \textsc{SmartPhone} fed the \textsc{TransPhoner} keyword into GPT-3 to automatically generate verbal cues and DALL·E to generate visual cues~\cite{lee2023smartphone}. \citet{lee2024overgenerate} introduced an overgenerate-and-rank approach, where LLMs overgenerate keyword sequences and verbal cues, and then rank them according to multiple different criteria.~\citet{balepur2024smart} aligned mnemonics with user preferences by fine-tuning Llama 2 for personalization and cost-efficiency. \citet{lee2025interpretable} learned latent user and Kanji (Chinese characters in Japanese) traits from a crowd-sourced platform for learning Kanji, and extract rules for constructing mnemonics using an Expectation-Maximization style algorithm.

These prior work focus predominantly on Indo-European L1-L2 language pairs with substantial phonological overlap~\cite{savva2014transphoner,lee2023smartphone}. However, typologically distant language pairs, such as English-Korean, present unique challenges that remain underexplored. English and Korean exhibit four major phonological mismatches that make mnemonic generation challenging. 
First, orthographic systems differ in dimensionality because English prints letters linearly, while Korean arranges the jamo into two-dimensional syllable blocks~\cite{Park_Li_2009}. 
Second, Korean forbids consonant clusters within a syllable, so epenthetic vowels must be inserted when adapting cluster-rich English words, which expands the syllable count~\cite{kang2003perceptual,Kenstowicz2005ThePA}. 
Third, certain English phonemes such as \textipa{/T/} have no direct Korean counterpart and are usually replaced with \textipa{/s/} or \textipa{/t/}~\cite{Tak2012Variation,Kim2011Lingua}. 
Fourth, the two languages exhibit phonemic contrast differences: Korean employs a three-way lenis, fortis, and aspirated stop contrast, whereas English distinguishes only voiced versus voiceless stops， and treats aspiration as a position-dependent allophone~\cite{Kang2014VOT,Cho2022Handbook}. 
These differences complicate the generation of phonologically faithful keyword sequence when Korean speakers learn English. 

\paragraph{Contributions} In this paper, we introduce a mnemonic generation system for language learning, \textsc{PhoniTale}. Our approach employs a greedy search through phonetically and syllabically approximated L2 sequences to identify the most suitable L1 keyword sequence. Specifically, we first transliterate L2 phonemes into L1-adapted sequences, segment these into syllables according to L1 phonological constraints, and then select keywords that maximize phonetic similarity while preserving syllabic structure. Unlike previous approaches that rely heavily on LLMs for keyword generation, we utilize LLMs only for verbal cue generation, while our specialized modules handle the cross-lingual phonological alignment. This design addresses the unique challenges posed by typologically distant languages, improves scalability, and mitigates hallucination risk. Through systematic evaluation including both automated metrics and human studies with short-term recall tests, we demonstrate that \textsc{PhoniTale} achieves comparable performance to human-authored mnemonics.

\section{Problem Statement}
\label{sec:problem_statement}

\begin{figure}
    \centering
    \captionsetup{skip=6pt}  
    \includegraphics[width=1.0\linewidth]{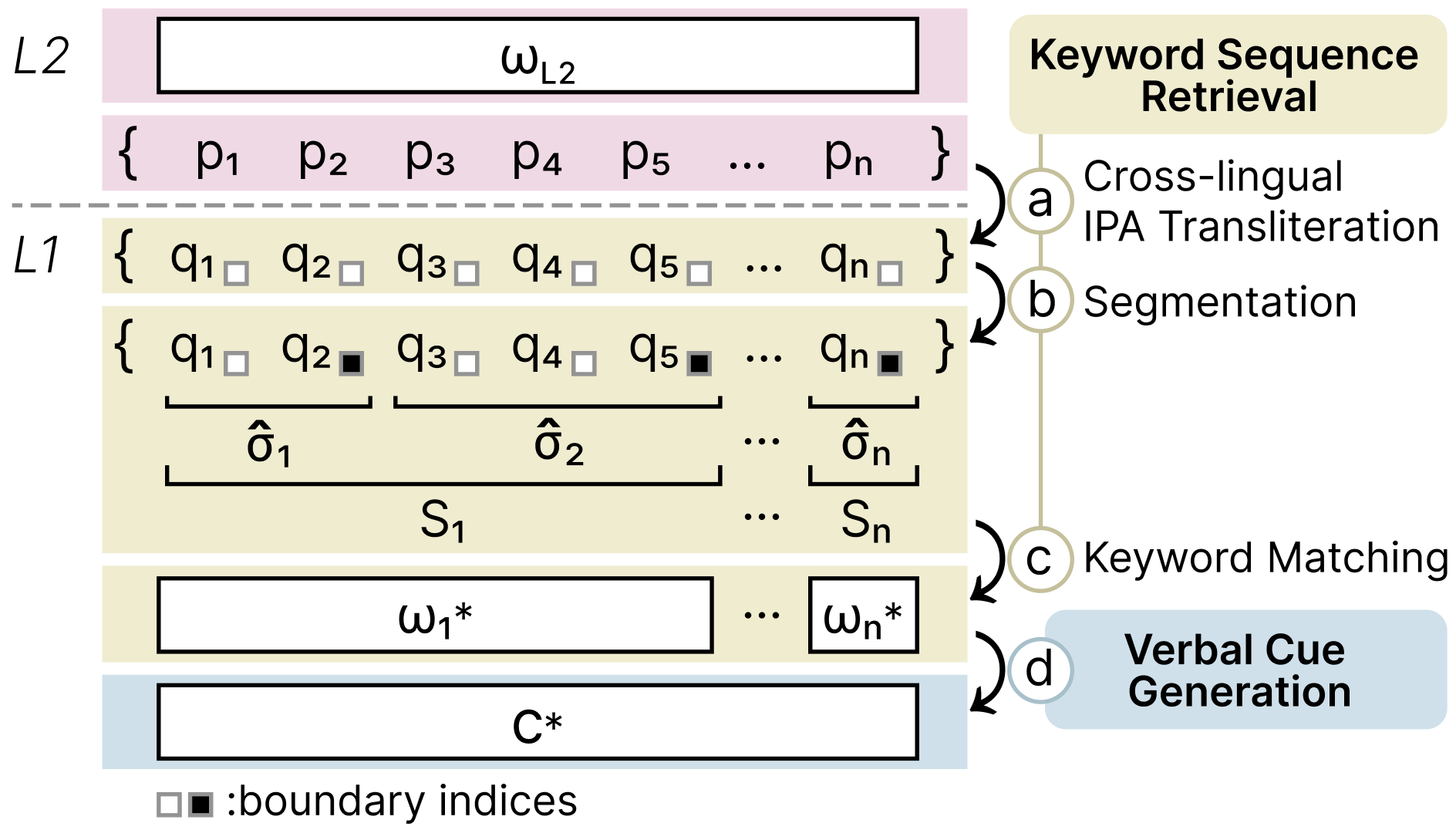}
    \caption{Problem formulation of the \textsc{PhoniTale} system. Phase~1, keyword sequence retrieval, comprises \textbf{(a)} IPA transliteration, \textbf{(b)} segmentation, and \textbf{(c)} keyword matching. Phase~2, \textbf{(d)}, performs verbal cue generation.}
   \label{fig:problem-statement}
\end{figure}

\textsc{PhoniTale} performs the task of retrieving cross-lingual phonologically similar keyword sequence and using them to construct a L1 verbal cue for a given L2 target word, following the process illustrated in Figure~\ref{fig:problem-statement}. Let $w_{\mathrm{L2}} \in \mathcal{V}_{\mathrm{L2}}$ be a word in the L2, and let $\ell$ denote its meaning. The goal is to retrieve a L1 keyword sequence $\mathcal{W}_{\mathrm{L1}} = (w_1^*, w_2^*, \ldots) \in \mathcal{V}_{\mathrm{L1}}$ that are phonologically similar to segments of $w_{\mathrm{L2}}$, and to use $\mathcal{W}_{\mathrm{L1}}$ to construct a verbal cue $c^* \in \mathcal{C}$, where $\mathcal{C}$ is the space of natural-language expressions in L1.

Retrieving phonologically similar keyword sequence begins by extracting the phoneme sequence of the L2 word, denoted $P_{\mathrm{L2}} = (p_1, p_2, \dots, p_m)$, where each $p_j \in \Sigma_{\mathrm{L2}}$, the L2 phoneme inventory. This sequence is then transliterated into an L1-adapted phoneme sequence $\widehat{P}_{\mathrm{L1}} = (q_1, q_2, \dots, q_n)$, with $q_i \in \Sigma_{\mathrm{L1}}$ as shown in Figure~\ref{fig:problem-statement}a, to approximate the L2 pronunciation using L1 phonological constraints.

Next, the adapted sequence is syllabified according to L1 phonological constraints. The syllabification process is non-deterministic, with multiple possible ways to divide the phoneme sequence. Each division creates different phoneme graphs that represent potential syllabification paths. From these multiple possibilities, a single path is selected that most closely aligns with L1 phonological patterns. This process yields the syllable sequence $\widehat{\boldsymbol{\sigma}} = (\sigma_1, \sigma_2, \ldots, \sigma_l)$.

These syllables are then grouped into $k$ segments $S_1, S_2, \dots, S_k$ using predefined partitioning rules. Each segment $S_i$ consists of one or more complete syllables and is defined by boundary indices $0 = b_0 < b_1 < \dots < b_k = l$ such that  $S_i = (\sigma_{b_{i-1}+1}, \sigma_{b_{i-1}+2}, \dots, \sigma_{b_i})$. The complete segmentation process is represented in Figure~\ref{fig:problem-statement}b.

Subsequently, each segment $S_i$ is then mapped to a keyword $w_i^* \in \mathcal{V}_{\mathrm{L1}}$ whose pronunciation closely resembles $S_i$ according to our phonological similarity criterion, as illustrated in Figure~\ref{fig:problem-statement}c. Finally, the verbal cue $c^*$ is generated by embedding the keyword sequence $\mathcal{W}_{\mathrm{L1}}$ in a natural-language expression that helps the learner associate the form of the L2 word with its meaning $\ell$, completing the process shown in Figure~\ref{fig:problem-statement}d. The complete output is the pair $(\mathcal{W}_{\mathrm{L1}}, c^*)$, which together support recall of the L2 word through phonological association.

\section{Methodology}

We divide the task into two phases: first, retrieving $\mathcal{W}_{\mathrm{L1}}$ for a given $w_{\mathrm{L2}}$ using our keyword sequence retrieval component; and second, using LLMs to generate verbal cues from $\mathcal{W}_{\mathrm{L1}}$, selecting the most coherent cue based on a ranking criterion.

\subsection{Keyword Sequence Retrieval}

We implement three modules by following the three steps of transforming $w_{\mathrm{L2}}$ into $\mathcal{W}_{\mathrm{L1}}$.

\subsubsection{Cross-lingual IPA Transliteration}

In the IPA transliteration module, we convert L2 phoneme sequence $P_{\mathrm{L2}}$ of $w_{\mathrm{L2}}$ into their L1-adapted sequence $P_{\mathrm{L1}}$. We utilize a neural sequence-to-sequence architecture with attention for the transduction task. We employ a bidirectional LSTM encoder and a unidirectional LSTM decoder, each with 256 hidden units~\cite{bahdanau2014neural}. The encoder processes $P_{\mathrm{L2}}$, capturing contextual information from both directions to produce a 512-dimensional representation, which the decoder uses to generate $\widehat{P}_{\mathrm{L1}}$.

We train the module by combining cross-entropy loss ($\mathcal{L}_{\text{CE}}$) and contrastive loss ($\mathcal{L}_{\text{cont}}$):
\begin{align}
    \mathcal{L}_{\text{CE}} &= - \sum_{t=1}^{T} \log P(y_t \mid y_{<t}, \mathbf{x}) \tag{2} \\
    \mathcal{L}_{\text{cont}} &= 1 - \frac{\mathbf{z}_{\text{enc}} \cdot \mathbf{z}_{\text{dec}}}{\norm{\mathbf{z}_{\text{enc}}}_2 \cdot \norm{\mathbf{z}_{\text{dec}}}_2} \tag{3}
\end{align}

The $\mathcal{L}_{\text{CE}}$ ensures token-level generation accuracy, while the $\mathcal{L}_{\text{cont}}$ promotes phonological consistency by aligning the encoder and decoder representations, $\mathbf{z}_{\text{enc}}$ and $\mathbf{z}_{\text{dec}}$, which are obtained by projecting their outputs into a shared embedding space via a lightweight feedforward layer. We combine these  losses as a weighted sum, $\mathcal{L}_{\text{total}} = \mathcal{L}_{\text{CE}} + \lambda \cdot \mathcal{L}_{\text{cont}}$, where we set $\lambda = 0.1$. We determined this weighting coefficient by analyzing the transliteration quality between $\widehat{P}_{\mathrm{L1}}$ and ground-truth L1-adapted phoneme sequence ${P}_{\mathrm{L1}}$ on our development set.

\subsubsection{Segmentation}
\label{sec:seg}
In our segmentation module, we predict $k$ contiguous phoneme segments $S$ from $\widehat{P}_{\mathrm{L1}}$ through a two-stage process. The first stage segments $\widehat{P}_{\mathrm{L1}}$ into a syllable sequence $\widehat{\boldsymbol{\sigma}} = (\sigma_1, \sigma_2, \ldots, \sigma_l)$, where each $\sigma_i \in \Sigma_{\mathrm{L1}}^+$ constitutes a valid L1 syllable. We assign binary labels to token positions in $P_{\mathrm{L1}}$ to indicate syllable boundaries, utilizing a bidirectional LSTM network with 256 hidden units in each direction. The network processes embedded IPA tokens augmented with binary vowel masks that signal potential syllabic nuclei~\cite{mayer2020syllable}.

The second stage combines these syllables into at most two segments $(S_1, S_2)$ for lengthy $w_{\mathrm{L2}}$ words, addressing the fundamental challenge that one-to-one mapping with $\mathcal{W}_{\mathrm{L1}}$ proves ineffective due to phoneme combinations or syllable structures in L2 that are absent in L1~\cite{daland-zuraw-2013-korean}. Each segment $S_i$ represents a contiguous sequence of L1-adapted syllables derived from the original L2 word.
For instance, L1-adapted phoneme sequence derived from the English word \emph{autopsy} might be syllabified as \textipa{/o/ - /t\textsuperscript{h}\textscripta p/ - /si/} and subsequently segmented into two segments such as $S_1$=\textipa{/ot\textsuperscript{h}\textscripta p/}  and $S_2$=\textipa{/si/}, or alternatively $S_1$=\textipa{/o/} and $S_2$=\textipa{/t\textsuperscript{h}\textscripta psi/.} This binary constraint prevents excessive fragmentation while preserving phonetic similarity and phonological coherence. The resulting segment sequence serves as input to the subsequent keyword sequence retrieval module, facilitating phonologically informed matching between $w_{\mathrm{L2}}$ and $\mathcal{W}_{\mathrm{L1}}$.

\subsubsection{Keyword Matching}

In the keyword matching module, we calculate phonological similarity between each segment $S_i$ and potential $\mathcal{W}_{\mathrm{L1}}$ from $\mathcal{V}_{\mathrm{L1}}$ to identify the most suitable matches. We convert phoneme sequences into 22-dimensional phonological feature vectors using PanPhon~\cite{mortensen2016panphon}, capturing distinctive phonological characteristics.
The similarity between a segment $S_i$ and a candidate keyword $\mathcal{W}_{\mathrm{L1},i}$ from Korean dictionary dataset~\cite{basic_korean_dict} is computed using the cosine similarity of their phonological feature embeddings with a structural alignment adjustment:
\begin{equation}
\phi(S_i, \mathcal{W}_{\mathrm{L1},i}) = \cos\left(\mathbf{v}(S_i), \mathbf{v}(\mathcal{W}_{\mathrm{L1},i})\right) + \Delta_{\text{structural}} \tag{4}
\end{equation}
where $\mathbf{v}(\cdot)$ represents the phonological feature embedding function and $\Delta_{\text{structural}}$ provides structural alignment adjustments.

While cosine similarity captures general phonological resemblance, this metric fails to account for syllable-level perception critical to Korean speakers~\cite{syllable_similarity, lee2017syllable, yoon2015perceptual,kang2003perceptual}. We incorporate four structural alignment adjustments in $\Delta_{\text{structural}}$: syllable overlap, initial-syllable match, early-phone alignment, and substring inclusion. Korean speakers perceive words as syllable bundles rather than phoneme strings, necessitating these adjustments to align our similarity function with native phonological perception processes. The initial-syllable match receives the highest weighting due to its greater perceptual significance in word recognition~\cite{lee2017syllable}.

Using the similarity function $\phi$, we pick the best keyword $\mathcal{W}_{\mathrm{L1},i}$ for each segment $S_i$ by
\begin{equation}
\mathcal{W}_{\mathrm{L1},i}^* = \arg\max_{w \in \mathcal{V}_{\mathrm{L1}}} \phi(S_i, w),
\tag{5}
\end{equation}
and score an entire segmentation by averaging each segment’s top match:
\begin{equation}
\frac{1}{m}\sum_{i=1}^{m}\max_{k \in \mathcal{V}_{\mathrm{L1}}}\mathrm{sim}(S_i, \mathcal{W}_{\mathrm{L1},i}).
\tag{6}
\end{equation}

This ranking process identifies the best keyword sequence $\mathcal{W}_{\mathrm{L1}} = (w_1^*, w_2^*)$ from our predefined segmentation candidates, selecting the keyword sequence that maximizes phonological similarity between the L2 word and the L1 keyword sequence. 

\subsection{Verbal Cue Generation}
\label{sec:verbal_cue_generation}
The verbal cue generation component builds upon \citet{lee2024overgenerate}, while introducing methodological refinements specific to the Korean-English language pair. We implement two major modifications to adapt the approach to a cross-lingual setting.

First, in the prompt, we eliminate the two-step approach used in \citet{lee2024overgenerate} which first generates a story and then summarizes that story to produce a verbal cue. While this approach aims to preserve keyword sequence in complex verbal cues, our cross-lingual setting with only two keywords, making this constraint unnecessary. We therefore directly generating without summarization which we validate through ablation studies presented in Section~\ref{sec:abl} (Appendix Table~\ref{tab:verbal_cue_prompt} for the prompt). 

Second, we discard the Age-of-Acquisition (AoA) ranking criterion from \citet{lee2024overgenerate} as it does not generalize effectively to cross-lingual contexts. The AoA of a word in L2 fails to reliably reflect its familiarity in L1. We retain only the context completeness criterion, calculating this by masking the target word in the verbal cue and prompting GPT-4o~\cite{openai2024gpt4o} to generate five probable candidates. We then compute the average cosine similarity between FastText~\cite{fasttext} embeddings of these candidates and the target word, trained on Korean corpus data~\cite{naver_shopping_corpus}. This approach quantifies how effectively the verbal cue provides context for learning the target word's meaning.

\section{Keyword Sequence Retrieval Validation}
In this section, we validate each module of our keyword sequence retrieval system. We first describe our datasets for training and evaluation, then present detailed results for each module.

\subsection{Dataset}
\label{sec:validation_dataset}

\paragraph{Keyword Pool} We construct a keyword candidate pool by filtering out non-lexical items such as grammatical particles, suffixes, and sentence-final endings from the Basic Korean Dictionary dataset~\cite{basic_korean_dict}.

The keyword pool consists of 55,316 unique entries that are phonologically representative and semantically well-formed.

\paragraph{Train} We construct a training dataset of 2,870 English–Korean word pairs with aligned IPA transcriptions. English vocabulary items originate from standardized GRE preparation materials, including official Educational Testing Service guides and commercial resources \cite{magoosh2021gre, princetonreview2020}. Each entry in our dataset includes an English word, its Korean transliteration, and IPA transcriptions for both languages, with syllable-level boundaries annotated in the Korean IPA.

For English, we obtain IPA transcriptions directly from the \citet{oedictionary}, which provides standardized phonetic representations widely used in linguistic research.

For Korean, we first extract transliterations of English words from the \citet{ahadictionary}. These transliterations are segmented into syllable blocks of the Korean writing system (Hangul), each composed of an initial consonant, a medial vowel, and an optional final consonant. Hangul is often characterized as an alphabetic syllabary, where individual graphemes (jamo) form syllabic blocks (kulja) corresponding to single phonological units \cite{pastore2019processing}. Each syllable block is then converted into its phonetic representation using the rule-based Hangul-to-IPA conversion method \cite{nam2021hangul}, which encodes standard Korean phonological processes \cite{shin2012phonological}. To identify syllable boundaries within the resulting IPA sequence, we extract the final IPA symbol from each block and annotate it with a binary indicator denoting syllable-final positions. This procedure enables consistent segmentation and alignment of syllable-level IPA representations across English and Korean.

\paragraph{Test} We use the book KSS~\cite{kyungsun2020} as our baseline for human-authored verbal cues, designed for native Korean speakers learning English. The vocabulary targets advanced-level standardized tests, including government employee entrance exams, university transfer admissions, the TOEFL~\cite{ets_toefl}, and the TEPS~\cite{teps_test}. From this vocabulary, we construct a test set of 36 words.

\subsection{Cross-lingual IPA Transliteration}
We validate the module using two metrics: Character Error Rate (CER) and Exact Match Rate (EMR). 
CER quantifies the proportion of character-level errors, including insertions, deletions, and substitutions, between the predicted output and the reference. This metric effectively captures fine-grained phonological discrepancies, which is important for transliteration tasks involving languages with complex phonotactics or lacking clear word boundaries. EMR measures the percentage of outputs that exactly match the reference sequences. It serves as a strict criterion for evaluating whether the model produces completely accurate transliterations. Our model achieves a CER of 3.95\% and an EMR of 75.56\% for the train data set.

To understand how our model addresses phonological divergence between English and Korean, we analyze attention patterns (See Appendix Figure~\ref{fig:attention_maps}). The visualizations reveal the model's strategies for cross-linguistic challenges: English affricates decompose into multiple Korean consonants; compatible sounds maintain one-to-one mappings; English diphthongs expand to accommodate Korean's vowel inventory; and syllable structures adapt to Korean phonotactic constraints. These patterns confirm the model's ability to dynamically adjust its mapping strategy based on input characteristics.

\subsection{Segmentation}
\begin{figure}[t]
   \centering
   \includegraphics[width=1.0\linewidth]{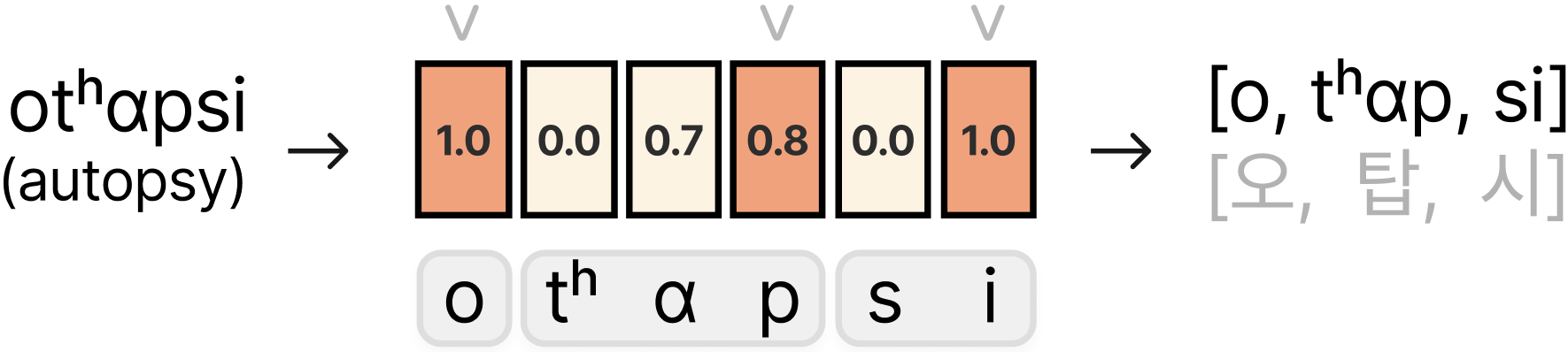}
   \caption{
       Visualization of the predicted syllable sequence of the English word \textit{autopsy}. For \textipa{/ot\textsuperscript{h}apsi/}, the model assigns high boundary probabilities after \textipa{o}, \textipa{p}, and \textipa{i}, segmenting the sequence into \textipa{[o, t\textsuperscript{h}ap, si]}.
   }
   \label{fig:syllable-prediction-example}
\end{figure}

We validate the module using boundary-level F1 score, which measures the model's precision and recall in identifying syllable boundaries. Since this module processes the output from the preceding transliteration component, we establish ground truth through manual annotation.

Figure~\ref{fig:syllable-prediction-example} illustrates the output of our model. For L1-adapted phoneme sequence \textipa{/ot\textsuperscript{h}\textscripta spsi/} derived from the English word \emph{autopsy}, the model assigns boundary probabilities that segment this sequence into phonologically valid Korean syllables.

Our model achieves a perfect boundary-level F1 score of 1.00 compared to reference boundaries on system transliteration outputs. This result indicates exceptional precision in identifying syllable junctures within the predicted L1-adapted sequence.

\subsection{Keyword Matching}
We validate the module by performing an ablation study on the structural alignment adjustment term to assess its contribution to our similarity function $\phi$. The experimental results confirm that this component significantly improves syllable-level matching, which constitutes a critical factor for Korean phonological perception~\cite{syllable_similarity}.

The comparison between retrieval methods shows clear differences in outcome quality. For the English word \emph{demolish} (IPA: \textipa{/dImalIS/}), the cosine similarity approach alone retrieves Korean keyword sequence with IPA transcriptions \textipa{/pimilli/} and \textipa{/Swi/}, whereas our complete similarity function identifies Korean keyword sequence transcribed as \textipa{/tEmullim/} and \textipa{/Swi/}. Similarly, for \emph{reckon} (IPA: \textipa{/rEk@n/}),  the cosine-only method produces Korean keyword sequence with IPA representations \textipa{/nE/} and \textipa{/k\super han/}, while our enhanced approach yields Korean keyword sequence represented as \textipa{/lEgE/} and \textipa{/k\super h2nsEp/}. These examples confirm that our weighting mechanism successfully prioritizes syllable matching as intended. 

The structural alignment adjustment enables our model to identify keyword sequence that preserve syllabic structure and phonological patterns aligned with Korean perceptual tendencies, even when this preservation necessitates selection of candidates with slightly greater overall phonological distance.
\section{Automated Evaluation}
\label{sec:auto-eval}

In the following sections, we detail how we evaluate the quality of the keyword sequence generated using \textsc{PhoniTale}, and we also discuss how we evaluate the quality of the generated verbal cues. 

\subsection{Dataset}
We use same dataset mentioned in \ref{sec:validation_dataset} for evaluation. We replicate \citet{lee2024overgenerate} in a same way for generating cross-lingual verbal cues, except that we translate prompts that were originally written in English with in-context examples for English-to-English learning into Korean, and use in-context examples from KSS to compare with \textsc{PhoniTale}. Hereafter, we refer to the human-authored book as \textbf{KSS}, \citet{lee2024overgenerate} as \textbf{OGR}, and \textsc{PhoniTale} as \textbf{PHT}. We use GPT-4o (temperature = 0.7) for both OGR and PHT throughout the entire pipeline to ensure fair comparison, following the temperature setting from \citet{lee2024overgenerate}.

\subsection{Metrics}

\subsubsection{Keyword Sequence}
We evaluate quality on three aspects: phonetic similarity, keyword omission, and keyword modification. We evaluate phonetic similarity using our IPA-based contrastive model, which measures how closely the concatenated Korean keyword sequence resembles the phonetic form of the English word.

We define keyword omission as the proportion of proposed keyword sequence that are missing from the generated verbal cue relative to the total keyword count. Since the mnemonic method depends on combining multiple keywords to approximate the target word, omitting even one can disrupt the intended phonetic connection.

We also track keyword modification, which represents the ratio of keywords that appear in altered forms relative to the total keyword count. These modifications can shift pronunciation away from the target word and weaken the mnemonic link. (See Appendix
Table \ref{tab:keyword_metric_examples} for examples.)

\subsubsection{Verbal Cue}
We evaluate the quality of verbal cue on \textit{context completeness} and \textit{perplexity} following \citet{lee2024overgenerate}. Again, as we do not use the imageability metric for keyword sequence, we also do not calculate the imageability score of the verbal cue.

We calculate context completeness as in Section~\ref{sec:verbal_cue_generation}, while we calculate perplexity as a proxy for coherence, using KoGPT2-base-v2~\cite{koGPT2}, OpenAI’s GPT-2 pretrained on large-scale Korean text data and adapted for natural language understanding and generation tasks in Korean.

\subsection{Results}
Table~\ref{tab:auto_eval} shows that PHT achieves superior performance compared to other methods in the evaluation of both keyword and verbal cue quality.

\begin{table*}[t]
\small
\centering
\begin{tabular}{cccc|cc}
\toprule
\textbf{Method} & \textbf{Phonetic$^\uparrow$} & \textbf{Omission$^\downarrow$}
& \textbf{Modification$^\downarrow$} & \textbf{Context$^\uparrow$} & \textbf{Perplexity$^\downarrow$}
 \\
\midrule
KSS & 0.74 & 3.7\% & - & 0.38 & 553.92\\
OGR & 0.86 & 3.4\% & 24.8\% &  0.29 & 691.01\\
PHT & \textbf{0.95} & \textbf{0\%} & \textbf{3.3\%} &  \textbf{0.39} & \textbf{433.41} \\
\bottomrule
\end{tabular}
\caption{Comparative analysis of metrics on keyword sequences and verbal cues. The keyword modifications of KSS was omitted because it does not provide information on the generation processes of keyword sequences and verbal cues.}
\label{tab:auto_eval}
\end{table*}

\subsubsection{Keyword Sequence}

OGR, relying on LLMs for generating keyword sequence, frequently includes keywords that either do not exist in standard lexicons or lack everyday usage frequency. This results in substantial modifications when the keyword sequence is converted to verbal cues. Further, the modifications reduce the phonetic similarity with target English words. 

KSS, authored by human, one possible reason for the low phonetic similarity is the its substitution of L1 meanings for L2 prefixes (e.g., \textit{re-}, \textit{in-}) and L2 suffixes (e.g., \textit{-cracy}).
For example, \textit{re-} is mapped to the L1 word meaning \textit{again}, and \textit{in-} is mapped to the L1 word meaning \textit{inside}.

\subsubsection{Verbal Cue}

OGR shows relatively lower performance in context completeness compared to other methods due to its excessive use of keywords. Since OGR focuses on splitting the target word into as many syllables as possible, the number of keywords corresponds to the number of syllables. Even with modified keywords not in standard lexicons or common use, it is generally difficult to generate natural and coherent context that effectively hints at the meaning of the target word. 

For example, for the target word \textit{frivolous}, OGR generates the keyword sequence \textipa{/p\textsuperscript{h}urwn/} (blue), \textipa{/pAl/} (field), and \textipa{/losw/} (Ross). The keyword sequence is shown in the verbal cue as ``The reckless woman who was scolded by \textit{Ross} in the \textit{blue} \textit{field},'' with perplexity score 689.3. On the other hand, PHT retrieves the keyword sequence \textipa{/p\textsuperscript{h}iri/} (flute) and \textipa{/palladw/} (Ballad), shown as: ``He, singing a \textit{ballad} with the \textit{flute}, acted rashly,'' with  perplexity  score of 231.0, which confirms that using two segments, as discussed in Section~\ref{sec:seg}, achieves better performance in cross-lingual setting.

Following the highest perplexity score observed with OGR, which results from the use of unconventional keywords, KSS shows the next highest score. This is most likely due to the incorporation of L2 morphological elements in the keywords, as mentioned earlier. These incorporation introduce irregularities that make the model harder to predict, resulting in higher perplexity scores. 
Beyond their surface inclusion, KSS often requires learners to disambiguate polysemous morphemes such as \textit{re-}, which can mean either \textit{again} or \textit{back} depending on the context. These inconsistencies in semantic interpretation and structural mapping increase irregularity in surface realizations, thereby hindering accurate verbal cue generation.

\subsection{Ablation Studies}
\label{sec:abl}
We conduct ablation studies on two key aspects: the prompts used for verbal cue generation and the models employed to generate these cues.
\subsubsection{Prompts}

\begin{table}[]
\small
\centering
\begin{tabular}{lcc}
\toprule
\textbf{Prompt} & \textbf{Context$^\uparrow$}
& \textbf{Perplexity$^\downarrow$}
\\
\midrule
OGR & 0.29 & 490.13 \\
PHT & \textbf{0.39} & \textbf{433.41} \\
\bottomrule
\end{tabular}
\caption{Comparison of verbal cue quality metrics using different prompt strategies while keeping all other components same as in the PHT system.}
\label{tab:ablation_prompt}
\end{table}

Table~\ref{tab:ablation_prompt} shows a result on ablating prompt for generating verbal cues. As discussed in Section~\ref{sec:verbal_cue_generation}, OGR uses two-step approach of generating a story then summarizing while PHT generates verbal cue right away. The result shows that PHT achieves better performance on both metrics, indicating that the two-step approach might be beneficial for English-only or verbal cue generations that require multiple keywords. However, in our cross-lingual setting, where there are only two keywords, generating verbal cue right away generates a better verbal cue.

\subsubsection{Language Model}

\begin{table}[]
\small
\centering
\begin{tabular}{lcc}
\toprule
\textbf{Model} & \textbf{Context$^\uparrow$} & \textbf{Perplexity$^\downarrow$} \\
\midrule
EXAONE3.5 & 0.38 & 450.30 \\
GPT-4o (PHT) & \textbf{0.39} & \textbf{433.41} \\
\bottomrule
\end{tabular}
\caption{Ablation results on verbal cue generation using different models.}
\label{tab:ablation_lm}
\end{table}

Table~\ref{tab:ablation_lm} shows a result on ablating language models for verbal cue generation. We utilize EXAONE3.5:32B, a 32-billion parameter open-sourced model with enhanced performance on Korean language tasks~\cite{research2024exaone35serieslarge} to test the language models suited for Korean language tasks can be an alternative for GPT4-o. The results shows that EXAONE3.5 achieves comparable performance on context completeness, and higher perplexity than GPT-4o. These results suggest that EXAONE-3.5 is a viable alternative, particularly when considering the cost and accessibility advantages of open-source models over proprietary ones.

\section{Human Evaluation}

\subsection{Participants}

We recruit Korean-native adults with intermediate English proficiency through university communities and LinkedIn. During the screening process, we also balance the participants’ proficiency levels across groups. After screening, we assign a total of 51 individuals, with 17 in each of the experimental groups: KSS, OGR, and PHT (see Appendix Section~\ref{appendix:participants} for details).

\subsection{Evaluation Setup}

We design our evaluation to jointly assess short-term recall~\cite{ellis1993psycholinguistic,savva2014transphoner,lee2023smartphone} and participant preference ratings~\cite{lee2024overgenerate}, to measure whether the verbal cues are helpful and whether learners prefer them. We implement a web platform to conduct an experiment comprising learning, testing, and feedback phases, where the learning phase differs across groups by presenting keyword sequences and verbal cues specific to each condition.

\subsection{Evaluation Procedure}

Participants complete three rounds of learning and testing, consisting of 12 English words. In the learning phase, they are presented with the English word, its Korean meaning, audio pronunciation, Korean keyword sequence, and a verbal cue. The testing includes two tasks: recognition (recalling the meaning of the English word) and generation (producing the English word). In the feedback phase, participants rate each verbal cue on three aspects on a 5-point Likert scale: \textit{helpfulness}, \textit{coherence}, and \textit{imageability} (See Appendix~\ref{appendix:procedure}  for procedure).

\subsection{Metrics}
\subsubsection{Correctness}
We assess the correctness of recognition and generation response using LLM-as-a-judge (GPT-4o)~\cite{chiang2023largelanguagemodelsalternative}. Previously, ~\citet{savva2014transphoner} employ Levenshtein distance for assessing correctness. However, as the responses from recognition might involve synonym usage, minor part-of-speech variations, or unintentional typos, relying solely on surface-level string similarity metrics like Levenshtein distance may lead to misleading evaluations. Therefore, we adopt a more semantically aware approach by leveraging GPT-4o as a judge to assess the alignment between the model output and the answer (See Appendix ~\ref{appendix:LLM-as-a-judge Prompt} for details).

\subsubsection{Preference Ratings}
We adopt the three criteria from \citet{lee2024overgenerate}, except that we replaced usefulness with helpfulness to assess how much each cue aided memorization, rather than measuring usefulness in the absence of a recall test. Helpfulness measures how effective the cue is for memorizing the English word. Coherence measures the logical soundness of the verbal cue. Imageability measures how well the cue evokes vivid imagery in the participant's mind.

\subsection{Results}

\subsubsection{Correctness}

\begin{figure}[h]
    \centering
    \includegraphics[width=0.95\linewidth]{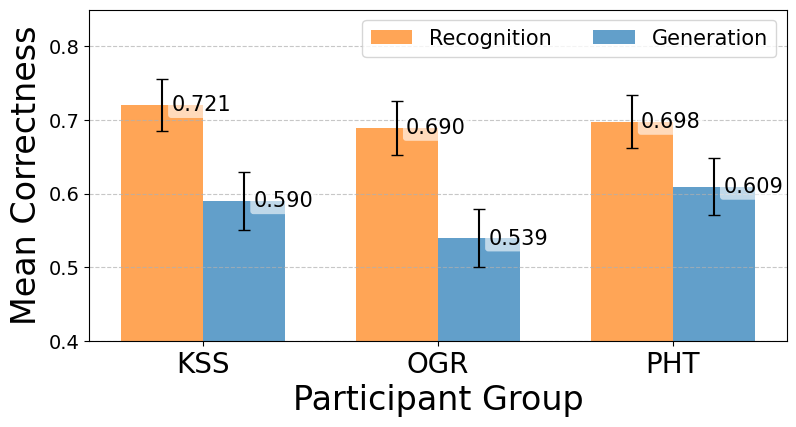}
    \caption{Mean correctness scores by participant group. Error bars indicate standard error.}
    \label{fig:correctness-bar}
\end{figure}

Figure~\ref{fig:correctness-bar} shows correctness scores for both recognition and generation tasks across groups. In the recognition task, KSS achieves the highest correctness, followed by PHT and OGR. However, statistical tests indicate no significant differences between the groups. In the generation task, \textsc{PHT} achieves the highest correctness, followed by KSS and OGR. Analysis shows that PHT significantly outperforms OGR ($p <.05$), while the difference between PHT and KSS is not statistically significant.

These results highlight two key points. First, PHT performs comparably to human-authored cues, suggesting that LLM-generated prompts can be as effective as those created by humans. Second, our focus on phonetic alignment, rather than imageability, proves beneficial in the context of Korean-English vocabulary learning.

\subsubsection{Preference Ratings}

\begin{table}[h]
    \centering
    \small
    \begin{tabular}{lccc}
    \toprule
    \textbf{Group} & \textbf{Helpfulness} & \textbf{Coherence} & \textbf{Imageability} \\
    \midrule
    \textsc{KSS} & 3.50 (1.41) & 3.64 (1.42) & 3.68 (1.39) \\
    \textsc{OGR} & 2.35 (1.40) & 2.33 (1.37) & 2.41 (1.41) \\
    \textsc{PHT} & 2.40 (1.26) & 2.26 (1.21) & 2.44 (1.23) \\
    \bottomrule
    \end{tabular}
    \caption{Comparison of mean (standard deviation) of 5-point Likert scale participant ratings by group.}
    \label{tab:survey-means}
\end{table}

Table \ref{tab:survey-means} shows preference ratings across groups. The KSS’s ratings are statistically significantly higher than the others across all three criteria ($p <.001$), indicating that participants prefer human-authored cues over LLM-generated ones.

PHT receives higher ratings than OGR for helpfulness and imageability, while OGR is rated higher for coherence. However, these differences were not statistically significant. Notably, although PHT achieves significantly higher correctness during generation, this does not correlate with helpfulness. This finding is consistent with prior work showing that subjective preference does not always align with verbal cue effectiveness~\cite{balepur2024smart}. In terms of coherence, OGR receives higher ratings because its cue generation transforms a meaningless keyword into a meaningful word, as shown in high modification rate, providing greater flexibility and resulting in more logically coherent cues.

\subsubsection{Case Study}

PHT achieves higher correctness than KSS by generating keyword sequences that better preserve consonantal structure while maintaining phoneme-level alignment with the target word. For example, in words containing \textipa{/r/} such as \textit{reckon} and \textit{render}, PHT selects initial keywords beginning with \textipa{/l/}, which is phonetically closer to \textipa{/r/}, whereas KSS selects \textipa{/n/}, resulting in less aligned mnemonics. We assume that keyword sequences with stronger phonological alignment contribute more effectively to learners’ ability to establish and retain accurate word associations.

However, KSS achieves higher correctness than PHT when keyword sequences are culturally rooted. For example, for \textit{felon}, KSS adapts the idiom ``to administer cudgel strokes'' into the verbal cue \textit{one who will cudgel-beat, therefore, a felon.} This construction is grammatically incorrect because it describes the one doing the beating rather than the one being beaten, yet learners readily reinterpret it as referring to the person who deserves punishment. The effectiveness of this vivid and culturally familiar cue is shown from its higher preference score compared to PHT. In contrast, PHT selects \textit{Peleus}, a mythological name that preserves phonological alignment but lacks cultural resonance, making it harder to remember. This case illustrates how KSS benefits from culturally rooted and expressive forms, while current LLMs, constrained by grammaticality, struggle to produce such non-standard yet pedagogically effective cues.

\section{Conclusion and Future Works}
In this paper, we introduced \textsc{PhoniTale}, a novel system combining keyword sequence retrieval with verbal cue generation. Automated and human evaluations show that our approach performs comparably to human-authored cues and outperforms the method proposed by \citet{lee2024overgenerate}. Furthermore, recall tests indicate our system achieves similar accuracy in recognition and statistically higher performance in generation. These results suggest that our strategy of leveraging phonetic similarity for mnemonic generation is effective.

Future extensions of this work are twofold. First, the system can be scaled to a broader range of typologically diverse languages, including syllable‐timed languages such as Japanese and Spanish and tonal languages such as Mandarin and Vietnamese. For example, in English–Japanese, the word \textit{render} pronounced as \textipa{/rEnd@r/} may be adapted as \textipa{/renda/} in Japanese, segmented into \textipa{[ren]} and \textipa{[da]}, and matched with native keywords such as レン ``ren'' (love) and だ ``da'' (copula) based on phonetic proximity. Provided that IPA‐aligned L2–L1 transliteration data is available, the framework adapted to new language pairs. Additional refinements, such as language‐specific syllable segmentation or tone modeling, can also enhance phonological compatibility across languages.

Second, we plan to extend the phoneme‐anchored retrieval system to code‐switched speech recognition. In such settings, phonological cues often transcend language boundaries, complicating identification of transliterated loanwords and domain-specific terms. By leveraging IPA-based representations, our approach offers a language-agnostic substrate for capturing cross-lingual phonetic similarity. This can improve recognition of borrowed or specialized vocabulary that deviates from canonical pronunciations, thereby reducing Word Error Rate (WER) in code-switched ASR scenarios. We plan to evaluate this by aligning phonetic units across typologically distant languages and assessing recognition gains for foreign-sounding or morphologically irregular tokens.

Together, these future directions aim to evolve PhoniTale into a more versatile, language‐agnostic tool with applications spanning multilingual mnemonic generation, resource creation, and speech recognition in phonologically diverse or code‐mixed environments.


\section*{Limitations}
Our investigation exhibits four primary constraints. First, we limit our research scope to English-Korean language pairs due to the limited availability of training data, necessitating future adaptations for other language combinations with distinct phonological structures and orthographic systems. Second, our evaluation methodology assesses only short-term recall performance rather than longitudinal retention. Future research requires delayed post-tests to evaluate long-term memory consolidation and mnemonic durability. Third, our vocabulary selection derives from standardized test materials targeting advanced-level English learners, potentially limiting \textsc{PhoniTale}'s applicability for beginning learners acquiring common vocabulary. Fourth, while our Korean dictionary dataset includes some conjugated forms, its lexical coverage remains limited. The absence of commonly used loanwords, neologisms, and other everyday variants reduces the pool of potential keywords and constrains the naturalness of generated verbal cues.

\section*{Acknowledgments}
This work was supported by the Institute of Information \& Communications Technology Planning \& Evaluation (IITP)-Global Data-X Leader HRD program grant funded by the Korea government (MSIT) (IITP-RS-2024-00440626)

\section*{Ethics Statement}
We conduct all human participant experiments in accordance with ethical guidelines after obtaining approval from our Institutional Review Board (IRB). Prior to participation, we present subjects with comprehensive consent forms detailing potential risks, anticipated benefits, expected time commitment, experimental activities, and compensation structure. All participants provide informed consent by acknowledging these terms before experiment commencement. The compensation structure aligns with local wage standards, with participants receiving between KRW 9,000 and 12,000 (approximately USD 6.50-8.70) based on the duration and complexity of their participation.


\bibliography{acl_latex}


\clearpage
\onecolumn           
\appendix            


\section*{\Large Appendix}


\section{Introduction}
\begin{figure*}[ht!]
    \centering
    \includegraphics[width=1\textwidth]{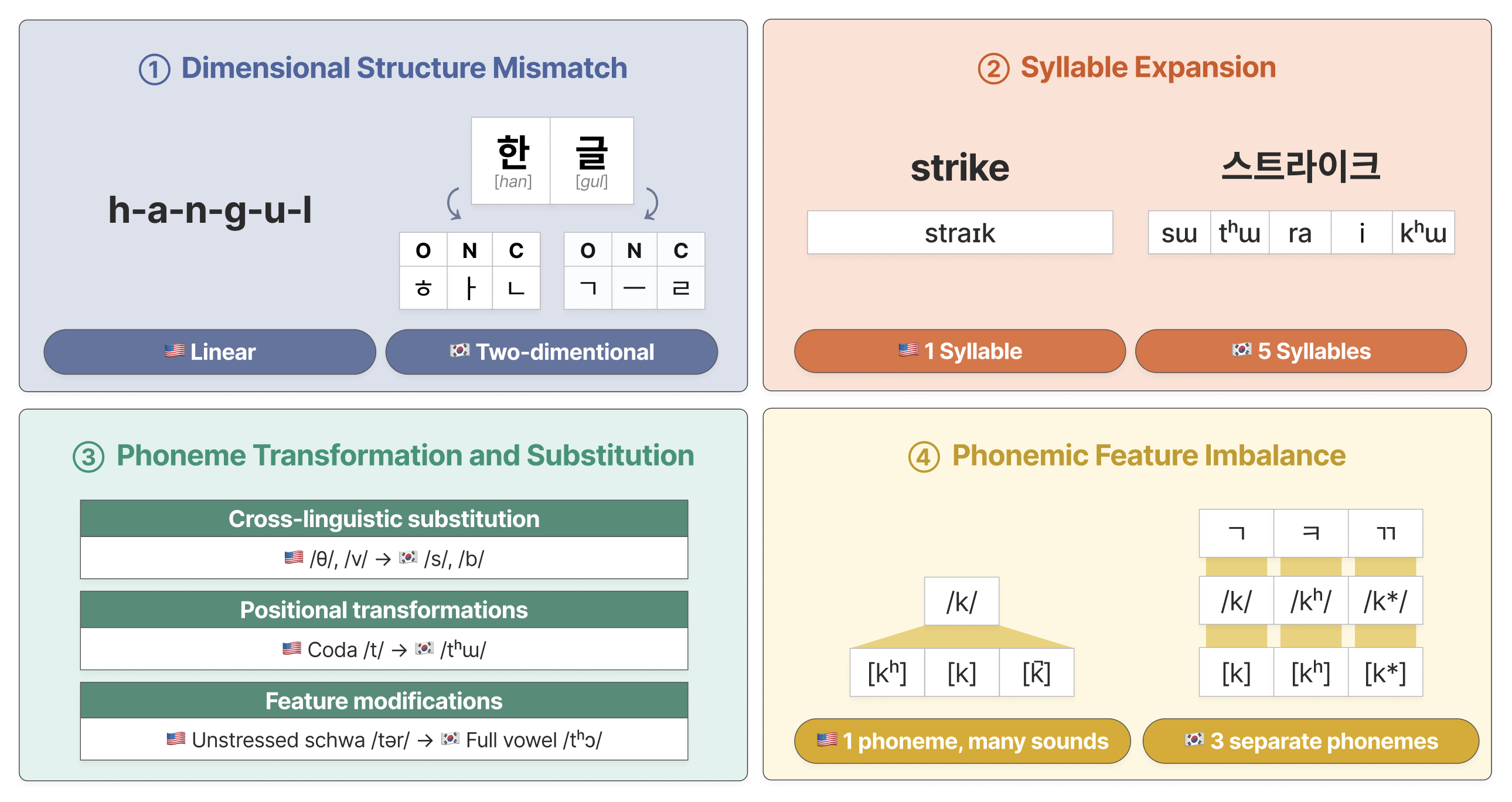}
    \caption{Four key challenges in English-Korean phonological alignment: (1) Dimensional structure mismatch: Korean's two-dimensional syllabic blocks versus English's linear sequence. (2) Syllable expansion due to consonant cluster resolution. (3) Phoneme transformation: Korean lacks certain English distinctions while English lacks Korean's three-way consonant contrast. (4) Phonemic contrast differences: Korean's systematic three-way distinction versus English's position-dependent allophones.}
    \label{fig:transliteration_challenge}
\end{figure*}


\section{Methodology}
\label{sec:methodology}

\subsection{\textsc{PhoniTale} Architecture}
\begin{minipage}{\textwidth}
   \centering
   \includegraphics[width=1\textwidth]{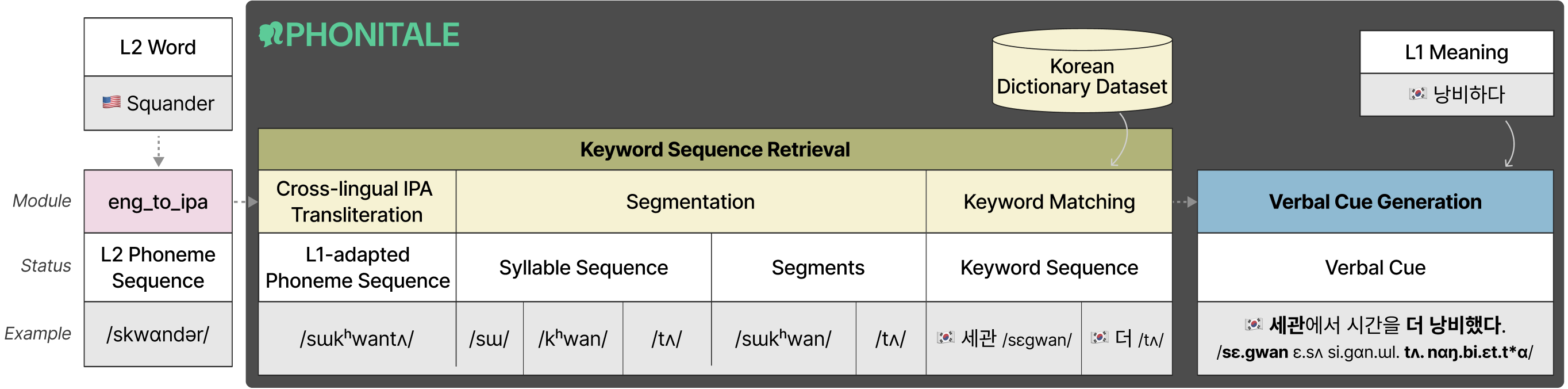}
   \captionof{figure}{
   We demonstrate this two-phase pipeline through a running example of $w_{\mathrm{L2}}$ ``squander''. The system first converts the word into $P_{\mathrm{L2}}$ (/\textipa{sk"wand\textreve r}/) using the \texttt{eng-to-ipa} library~\cite{eng_to_ipa}, which is based on the CMU Pronouncing Dictionary~\cite{cmu_dictionary}. The system then generates $\widehat{P}_{\mathrm{L1}} $ (/\textipa{sWk\super{h}want\textturnv}/), predicts syllable sequence (/\textipa{sW}/, /\textipa{k\super{h}wan}/, /\textipa{t\textturnv}/), and derives the segments (/\textipa{sWk\super{h}wan}/, /\textipa{t\textturnv}/). The system retrieves $\mathcal{W}_{\mathrm{L1}}$ with IPA transcriptions /\textipa{s\ae .gwan}/ and /\textipa{t\textturnv}/, and uses them to construct a verbal cue: ``\textipa{\textbf{sE.gwan} E.s\textturnv\tipamedspace si.gan.\textomega l. \textbf{t\textturnv} \textbf{naN.bi.Et.t*a}}'' (English translation: \textbf{Wasted} \textbf{more} time at \textbf{customs}).}
   \label{fig:methdology_architecture}
\end{minipage}


\subsection{Syllable Prediction} 
\label{app:syllable-prediction}
\begin{minipage}{\textwidth}
   \centering
   \includegraphics[width=0.48\textwidth]{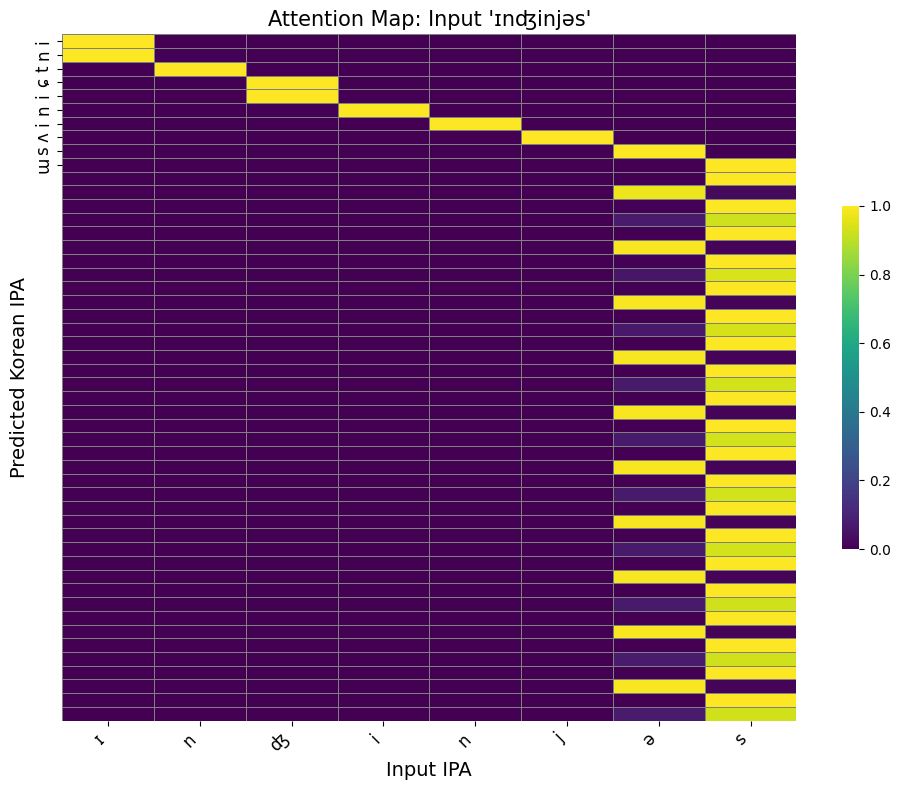}%
   \hfill
   \includegraphics[width=0.48\textwidth]{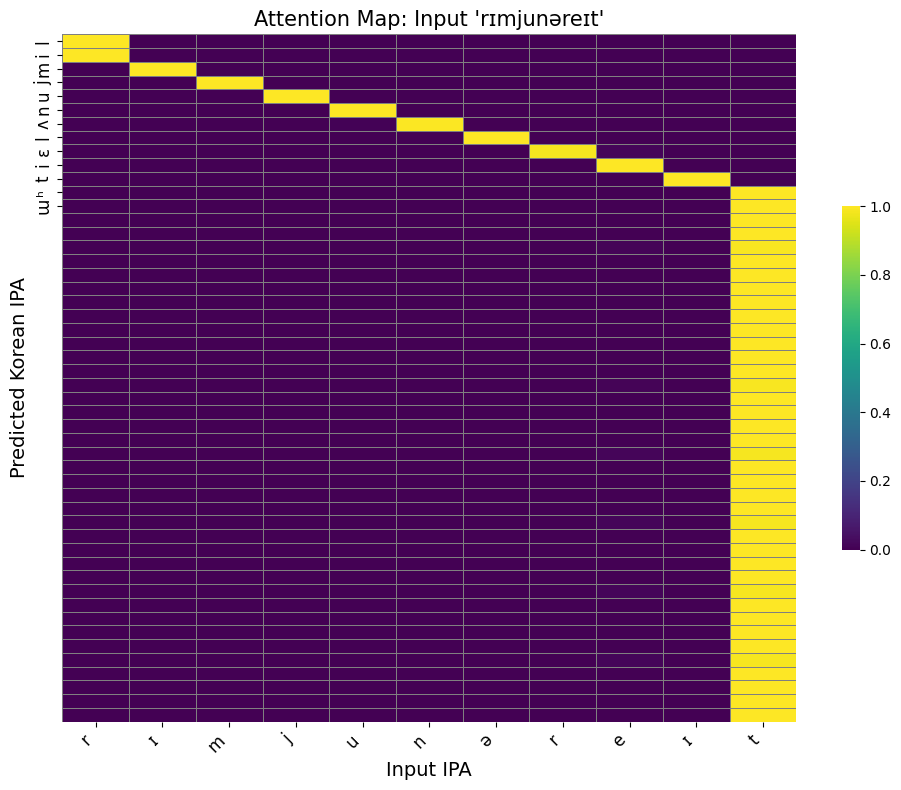}
   \vspace{0.8em}
   \includegraphics[width=0.48\textwidth]{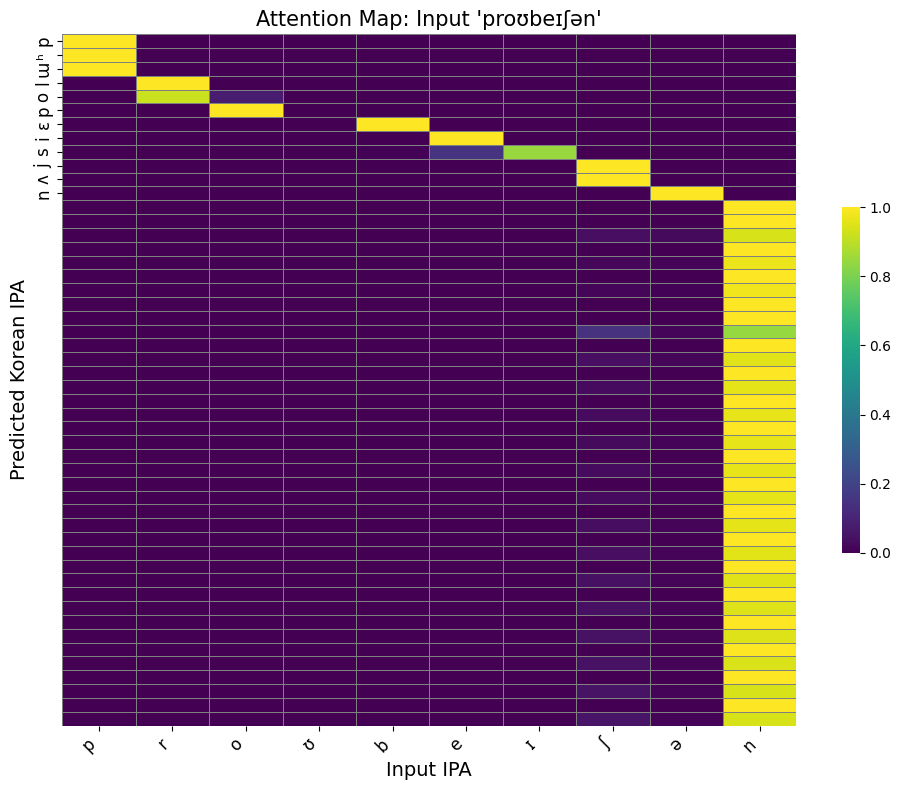}%
   \hfill
   \includegraphics[width=0.48\textwidth]{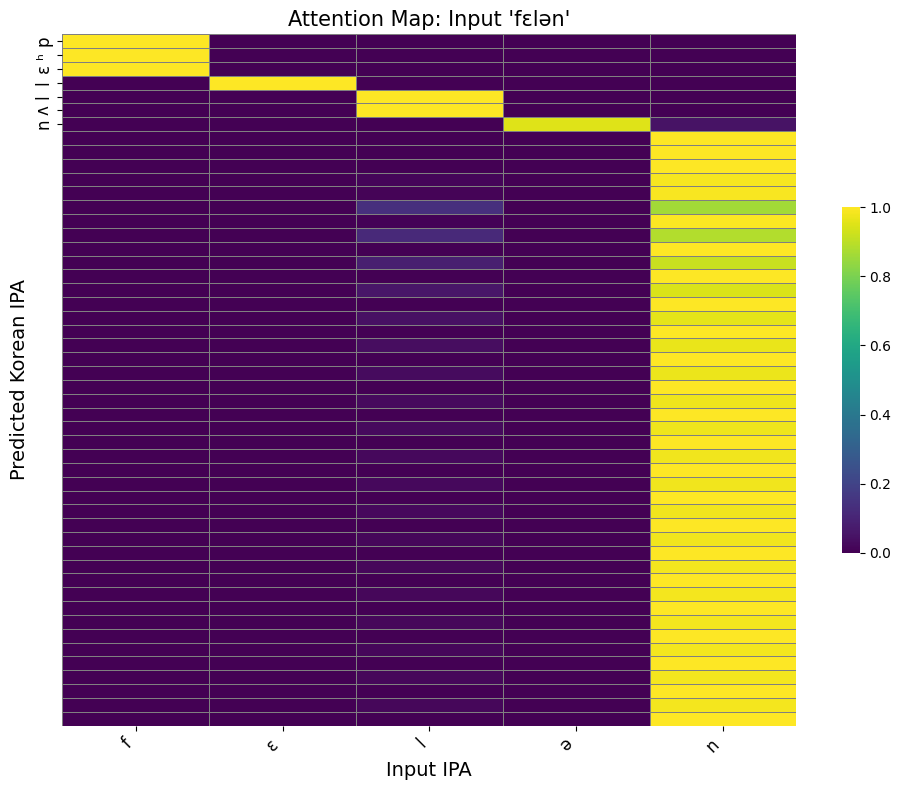}
   \captionof{figure}{
       Representative attention maps visualizing cross-lingual phonological alignment patterns. \textbf{Top left}: Attention map for \textipa{/IndIdZ@ni@s/} (indigenous) showing affricate decomposition where English affricates are mapped to multiple Korean consonants, creating vertical attention patterns. \textbf{Top right}: Attention map for \textipa{/rImjun@reIt/} (remunerate) demonstrating near one-to-one alignment with strong diagonal attention patterns where phonological structures are compatible across languages. \textbf{Bottom left}: Attention map for \textipa{/proUbeIS@n/} (probation) highlighting diphthong expansion and coda realignment where English diphthongs spread across multiple Korean vowels and final consonants shift to match Korean syllable constraints. \textbf{Bottom right}: Attention map for \textipa{/fEl@n/} (felon) illustrating structure-induced elongation where the short CVC English syllables must expand to fit Korean's more restrictive syllable templates, creating distributed attention across additional segments.}
   \label{fig:attention_maps}
\end{minipage}

\subsection{Keyword Matching} 
\label{app:methodology-algorithm}
\begin{minipage}{\textwidth}
\noindent\rule{\textwidth}{0.4pt} 
\begin{algorithmic}[1]
\Require Segment $S_i$, Candidate keyword $w^*_{\mathrm{L1},i}$
\Require Embedding function $\mathbf{v}(\cdot)$
\Require Parameters: $\lambda_{\text{syll}}$, $\lambda_{\text{first}}$, $\lambda_{\text{substr}}$, $\lambda_{\text{early}}$
\State $score \gets \cos\left( \mathbf{v}(S_i), \mathbf{v}(w^*_{\mathrm{L1},i}) \right)$ \Comment{Base similarity}
\If{\Call{SyllableOverlap}{$S_i$, $w^*_{\mathrm{L1},i}$}} 
    \If{\Call{InitialSyllableMatch}{$S_i$, $w^*_{\mathrm{L1},i}$}}
        \State $score \gets score + \lambda_{\text{syll}} \times \lambda_{\text{first}}$
    \Else
        \State $score \gets score + \lambda_{\text{syll}}$
    \EndIf
\Else
    \If{\Call{SubstringInclusion}{$S_i$, $w^*_{\mathrm{L1},i}$}} 
        \State $score \gets score + \lambda_{\text{substr}}$
    \ElsIf{\Call{EarlyPhoneMatch}{$S_i$, $w^*_{\mathrm{L1},i}$}} 
        \State $score \gets score + \lambda_{\text{early}}$
    \EndIf
\EndIf
\State \Return $score$
\end{algorithmic}
\vspace{0.5em}
\noindent\rule{\textwidth}{0.4pt}
\vspace{0.3em}
\captionof{algorithm}{
Keyword scoring algorithm used in the keyword matching module.
We apply cosine similarity as the base score, and augment it with structural alignment adjustments, which are empirically tuned on the development set:
$\lambda_{\text{syll}} = 0.9$ (syllable overlap bonus),
$\lambda_{\text{first}} = 2.0$ (initial-syllable match multiplier),
$\lambda_{\text{substr}} = 0.3$ (substring inclusion bonus),
and $\lambda_{\text{early}} = 0.2$ (early-phone match bonus).
These additions are designed to enhance alignment with syllable-based perception patterns in Korean, aiding memorability and cue effectiveness.
}
\label{alg:keyword-matching}
\end{minipage}

\subsection{\textsc{PhoniTale} Prompt}
\label{sec:prompt}

The prompt was originally designed in Korean. For reproducibility, we provide both the original and its English translation.

\renewcommand{\arraystretch}{1.3}

\begin{longtable}{|p{2cm}|p{12cm}|}
\hline
Prompt & 게임 이름: 이야기 엮기 놀이 \\
 & \small\textit{Game name: Story-Chaining Game}\\
 & \\
 & 게임 설명: 이야기 엮기 놀이에서 플레이어들은 목표 단어 후보와 키워드 세트를 받습니다. 플레이어들의 임무는 이 단어들을 교묘하게 사용하여 짧고 간결한 한 문장의 이야기를 만드는 것입니다. 궁극적인 도전은 목표 단어 후보 중 하나를 선택적으로 포함하고 제시된 순서대로 정확히 키워드를 포함하는 한 문장의 이야기를 구성하는 것입니다. \\
 & \small\textit{Game description: In the Story-Chaining Game, players receive a target word candidate set and a keyword set. Their task is to craft a short, concise, single-sentence story that cleverly incorporates these words. The ultimate challenge is to construct a sentence that includes at least one of the target word candidates of your choice and strictly uses the provided keywords in the given order.} \\
 & \\
 & 게임 규칙/제약사항: \\
 & \small\textit{Game rules/constraints:} \\
 & 1. 각 플레이어는 목표 단어 후보와 키워드 세트를 받습니다. \\
 & \small\textit{1. Each player receives a set of target word candidates and keywords.} \\
 & 2. 목표 단어를 먼저 결정해야 합니다. 목표 단어 후보가 하나라면, 그 단어가 곧 목표 단어가 됩니다. 목표 단어 후보가 여러 개라면, 플레이어는 그 중 하나를 선택해야 합니다. \\
 & \small\textit{2. The target word must be chosen first. If there is only one candidate, that word becomes the target word. If multiple candidates are provided, the player must select one.} \\
 & 3. 목표 단어와 키워드를 사용하여 한 문장으로 된 짧은 이야기를 만들어야 합니다. \\
 & \small\textit{3. The target word and keywords must be used to create a short, single-sentence story.} \\
 & 4. 키워드는 주어진 순서대로 정확히 등장해야 합니다. \\
 & \small\textit{4. Keywords must appear exactly in the specified order.} \\
 & 5. 한 문장의 이야기에는 목표 단어가 포함되어야 하며, 한 번만 나타나야 합니다. 목표 단어는 꺾쇠 괄호(< >)로 묶어 강조해야 합니다. \\
 & \small\textit{5. The story must include the target word exactly once, and it should be highlighted using angle brackets (< >).} \\
 & 6. 전체 내용은 json 형식으로 반환해야 합니다. \\
 & \small\textit{6. The entire output must be returned in JSON format.} \\
 & 7. 플레이어는 키워드의 순서를 재배열하는 것이 엄격히 금지됩니다. \\
 & \small\textit{7. Rearranging the order of the keywords is strictly prohibited.} \\
 & \\
 & 다음은 입력과 출력이 어떻게 보여야 하는지에 대한 예시입니다: \\
 & \small\textit{The following are examples of the expected input and output format:} \\
 & \\
 & [Input] \\
 & 목표 단어 후보: <취소하다> \\
 & \small\textit{Target word candidates: <countermand>} \\
 & 키워드 세트: 카운터, 만두 \\
 & \small\textit{Keyword set: /\textipa{k\super{h}a.un.t\super{h}\textturnv}/ (counter), /\textipa{man.du}/ (dumpling)} \\
 & [Output] \\
 & \{ \\
 & \ \ \ \ "목표 단어": "취소하다", \\
 & \ \ \ \ \small\textit{"target word": "countermand",} \\
 & \ \ \ \ "이야기": "그는 카운터에서 만두 주문만 <취소했다>." \\
 & \ \ \ \ \small\textit{"story": "He <countermanded> the dumpling order at the counter."} \\

 & \} \\
 & [Input] \\
 & 목표 단어 후보: <범인, 범죄자> \\
 & \small\textit{Target word candidates: <culprit, criminal>} \\
 & 키워드 세트: 칼, 뿌리다 \\
 & \small\textit{Keyword set: /\textipa{k\super{h}al}/ (knife), /\textipa{[p\textsuperscript{h}u.\textit{r}i.da]}/ (scatter)} \\
 & [Output] \\
 & \{ \\
 & \ \ \ \ "목표 단어": "범죄자", \\
 & \ \ \ \ \small\textit{"target word": "culprit",} \\
 & \ \ \ \ "이야기": "칼을 뿌리는 <범죄자>." \\
 & \ \ \ \ \small\textit{"story": "The <culprit> scattered knives."} \\
 & \} \\
 & [Input] \\
 & 목표 단어 후보: <튀기다, 첨벙거리다> \\
 & \small\textit{Target word candidates: <fry, splash>} \\
 & 키워드 세트: 수풀, 쉬 \\
 & \small\textit{Keyword set: /\textipa{su.p\super{h}ul}/ (bush), /\textipa{\c{c}i}/ (pee)} \\
 & [Output] \\
 & \{ \\
 & \ \ \ \ "목표 단어": "튀기다", \\
 & \ \ \ \ \small\textit{"target word": "splash",} \\
 & \ \ \ \ "이야기": "수풀에 쉬를 하다 물을 <튀겼다>." \\
 & \ \ \ \ \small\textit{"story": "While peeing in the bush, water <splashed>."} \\
 & \} \\
 & [Input] \\
 & 목표 단어 후보: <나태한, 게으른> \\
 & \small\textit{Target word candidates: <sluggish, lazy>} \\
 & 키워드 세트: 인어, 덜렁대다 \\
 & \small\textit{Keyword set: /\textipa{in.\textturnv}/(mermaid), /\textipa{t\textturnv l.l\textturnv \ng .d\textepsilon.da}/ (fumble)} \\
 & [Output] \\
\hline
Response 
 & \{ \\
 & \ \ \ \ "목표 단어": "게으른", \\
 & \ \ \ \ \small\textit{"target word": "indolent",} \\
 & \ \ \ \ "이야기": "인어는 덜렁대며 <게으르게> 움직였다." \\
 & \ \ \ \ \small\textit{"story": "The mermaid fumbled around and moved indolently."} \\
 & \} \\
\hline
\caption{Prompts for generating verbal cues.} \label{tab:verbal_cue_prompt} \\
\end{longtable}

\section{Automatic Evaluation}

\begin{minipage}{\textwidth}
\centering
\begin{tabular}{p{2cm} p{2cm} p{2.4cm} p{2.4cm} p{5cm}}
    \toprule
    \textbf{Issue Type} & \textbf{Target Word} & \textbf{Proposed \newline Keyword \newline Sequence} & \textbf{Used \newline Keyword \newline Sequence} & \textbf{Description} \\
    \midrule
    Omission & provisional & \textipa{p\textsuperscript{h}wro}, \textipa{pis2}, \textipa{n2l} & \textipa{p\textsuperscript{h}wro}, \textipa{pis2} & 
    The keyword \textipa{n2l} is omitted from the verbal cue. \\
    \addlinespace
    Modification & reticent & \textipa{lE}, \textipa{t\textsuperscript{h}i}, \textipa{sEnt\textsuperscript{h}w} & \textipa{lEswt\textsuperscript{h}oraN}, \textipa{t\textsuperscript{h}i}, \textipa{sEnt\textsuperscript{h}w} & 
    The keyword \textipa{lE} is modified to \textipa{l@swt\textsuperscript{h}oraN}. \\
    \bottomrule
\end{tabular}
\captionof{table}{Examples of Keyword Omission and Modification}
\label{tab:keyword_metric_examples}
\end{minipage}


\section{Human Evaluation}

\subsection{Participants}
\label{appendix:participants}

\begin{minipage}{\textwidth}
We summarize the participant recruitment, screening, and group assignment process in Table~\ref{tab:participant-summary}. Vocabulary familiarity was assessed using a 12-word survey. Words were grouped into difficulty tiers and scored (3=High, 2=Medium, 1=Low) based on the percentage of participants who reported familiarity (see Table~\ref{tab:appendix-words-familiarity}). Final group assignment ensured balance in vocabulary familiarity, age, and education level.
\end{minipage}

\vspace{1em}

\begin{minipage}{\textwidth}
\centering
\renewcommand{\arraystretch}{1.4}
\scalebox{0.95}{
\begin{tabular}{p{4cm} p{11cm}}
\toprule
\textbf{Step} & \textbf{Description} \\
\midrule
Recruitment & 167 Korean-native adults via university communities and LinkedIn \\
\midrule
Screening Task & 12-word self-report survey \newline (SAT/TOEFL/GRE vocabulary) \\
\midrule
Scoring Method & Score = sum of recognized words weighted by difficulty \newline (3 = high, 2 = medium, 1 = low) \\
\midrule
Filtering & Top outliers removed using upper quartile threshold; \newline bottom excluded if score ≤ 2 \newline  → 132 eligible participants \\
\midrule
Group Assignment & Random assignment to \textsc{KSS}, \textsc{OGR}, and \textsc{PHT} with matched familiarity levels \\
\midrule
Experiment Completion & 55 participants completed the main task \\
\midrule
Quality Filtering & Bottom 4 participants excluded based on lowest completeness scores \\
\midrule
Final Sample & 51 participants (17 per group) \\
\midrule
Equivalence Check & No significant differences across groups:
\begin{itemize}
  \item Vocabulary familiarity ($p = .4378$)
  \item Age ($p = .9100$)
  \item Education level ($p = .3599$)
\end{itemize} \\
\bottomrule
\end{tabular}
}

\captionof{table}{Summary of participant recruitment, screening, and group assignment process.}
\label{tab:participant-summary}
\end{minipage}

\vspace{2em}

\begin{minipage}{\textwidth}
\centering
\renewcommand{\arraystretch}{1.2}
\begin{tabular}{lcccc}
\toprule
\textbf{Group} & \textbf{Mean Age} & \textbf{Std Dev} & \textbf{Min Age} & \textbf{Max Age} \\
\midrule
KSS & 28.24 & 6.47 & 20 & 41 \\
OGR & 28.41 & 8.54 & 18 & 54 \\
PHT & 28.47 & 5.56 & 18 & 35 \\
\bottomrule
\end{tabular}
\captionof{table}{Age statistics by group}
\label{tab:age-stats}
\end{minipage}

\vspace{2em}

\begin{minipage}{\textwidth}
\centering
\renewcommand{\arraystretch}{1.2}
\scalebox{1}{
\begin{tabular}{>{\raggedright}m{3.5cm} >{\raggedright}m{3.5cm} c r}
\toprule
\textbf{Target Word} & \textbf{Difficulty Level} & \textbf{Assigned Score} & \textbf{Familiarity} \\
\midrule
intransigent & High   & \textbf{3} & 1.8\%  \\
reinstate    & High   & \textbf{3} & 6.0\%  \\
horrendous   & High   & \textbf{3} & 13.8\% \\
sanction     & High   & \textbf{3} & 21.0\% \\
abdominal    & Medium & \textbf{2} & 25.1\% \\
uphold       & Medium & \textbf{2} & 38.9\% \\
deduce       & Medium & \textbf{2} & 42.5\% \\
mutable      & Medium & \textbf{2} & 46.7\% \\
hygiene      & Low    & \textbf{1} & 54.5\% \\
criterion    & Low    & \textbf{1} & 55.7\% \\
inverse      & Low    & \textbf{1} & 78.4\% \\
align        & Low    & \textbf{1} & 81.4\% \\
\bottomrule
\end{tabular}
}
\captionof{table}{Twelve English words used in the screening survey, grouped by assigned difficulty level \\ and annotated with the percentage of participants who reported familiarity. These values serve \\ as the basis for scoring vocabulary familiarity across participants.}
\label{tab:appendix-words-familiarity}
\end{minipage}

\subsection{Evaluation Procedure}
\label{appendix:procedure}

\setlength{\tabcolsep}{15pt}
\renewcommand{\arraystretch}{1.0}

\subsubsection{Procedure}

\begin{longtable}{|p{2.5cm}|p{11cm}|}

\hline
\textbf{Phase} & \textbf{Description} \\
\hline
\endfirsthead

\endhead

\hline
\textbf{Learning} &
\begin{minipage}[t]{\linewidth}\vspace{0.5em}
\begin{itemize}
  \item For each word, participants see:
  \begin{itemize}
    \item English word (visually segmented)
    \item Korean definition
    \item Audio pronunciation (played at 2s and 7s)
    \item Korean keyword sequence
    \item Verbal cue
  \end{itemize}
  \item Color underlining highlights phonological alignment
  \item 30-second time limit (advance allowed after 15s)
  \item 1-second blank screen between words
\end{itemize}
\vspace{0.5em}\end{minipage}
\\ \hline

\textbf{Testing – Recognition} &
\begin{minipage}[t]{\linewidth}\vspace{0.5em}
\begin{itemize}
  \item Task: Type the Korean definition
  \item For each word:
  \begin{itemize}
    \item English word
    \item Audio pronunciation (played at 2s and 7s)
  \end{itemize}
  \item 30-second time limit
  \item 1-second blank screen between words
  \item Responses are used to compute correctness scores
\end{itemize}
\vspace{0.5em}\end{minipage}
\\ \hline

\textbf{Testing – Generation} &
\begin{minipage}[t]{\linewidth}\vspace{0.5em}
\begin{itemize}
  \item Task: Type the English word
  \item For each word:
  \begin{itemize}
    \item Korean definition
  \end{itemize}
  \item 30-second time limit
  \item 1-second blank screen between words
  \item Responses are used to compute correctness scores
\end{itemize}
\vspace{0.5em}\end{minipage}
\\ \hline

\textbf{Feedback} &
\begin{minipage}[t]{\linewidth}\vspace{0.5em}
\begin{itemize}
  \item Participants rate each mnemonic on three 5-point Likert scales:
  \begin{itemize}
    \item Helpfulness: Cue supports recall of the word’s meaning
    \item Coherence: Sentence is logical and grammatically natural
    \item Imageability: Cue evokes a vivid and concrete image
  \end{itemize}
\end{itemize}
\vspace{0.5em}\end{minipage}
\\ \hline
\caption{Detailed task structure for learning, testing, and feedback phases. Each round includes 12 English words, repeated across three rounds (36 total).}
\label{tab:appendix-procedure} \\

\end{longtable}

\subsubsection{Web User Interface} 

\begin{minipage}{\textwidth}
    \centering
    \includegraphics[width=1.0\textwidth]{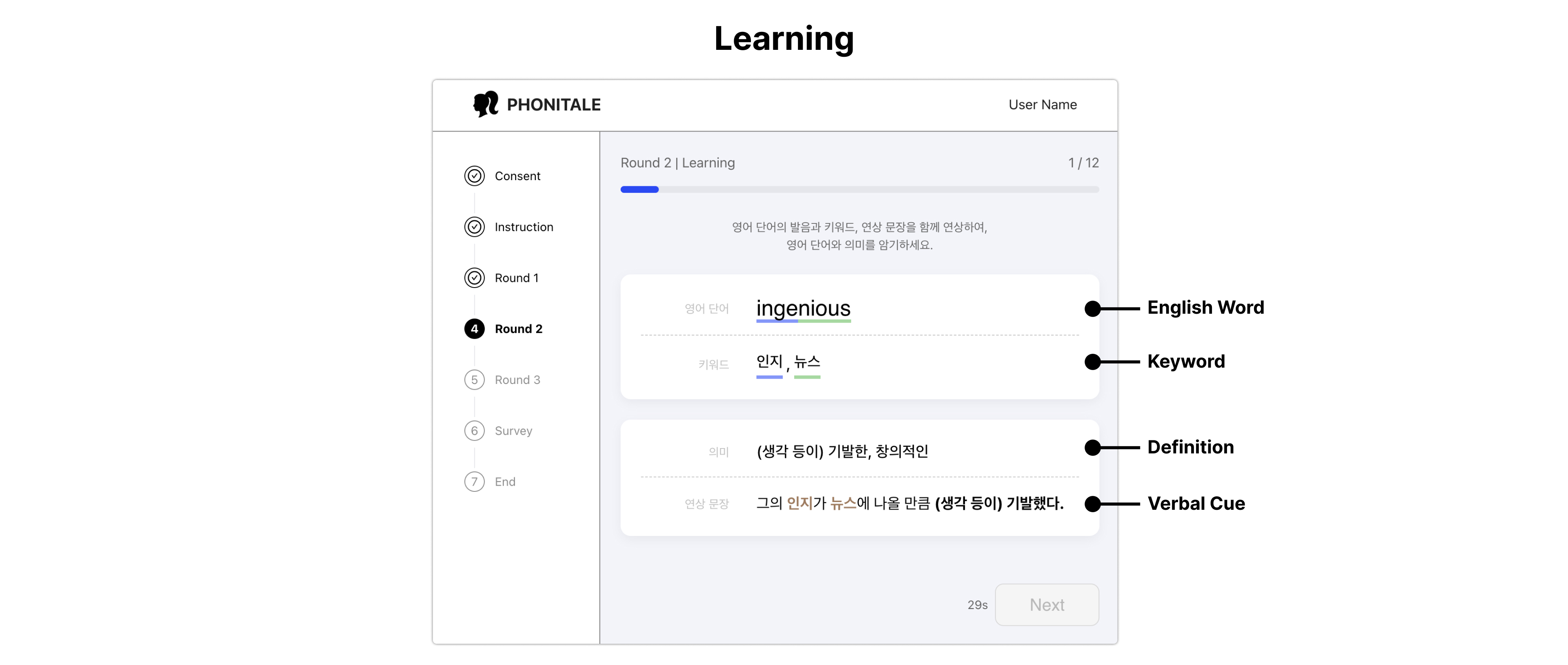}
    \captionof{figure}{User interface for the learning phase. Each screen presents the English word, Korean definition, phonologically aligned keyword sequence, and a verbal cue.}
    \label{fig:ui-learning}
\end{minipage}

\begin{minipage}{\textwidth}
    \centering
    \includegraphics[width=0.9\textwidth]{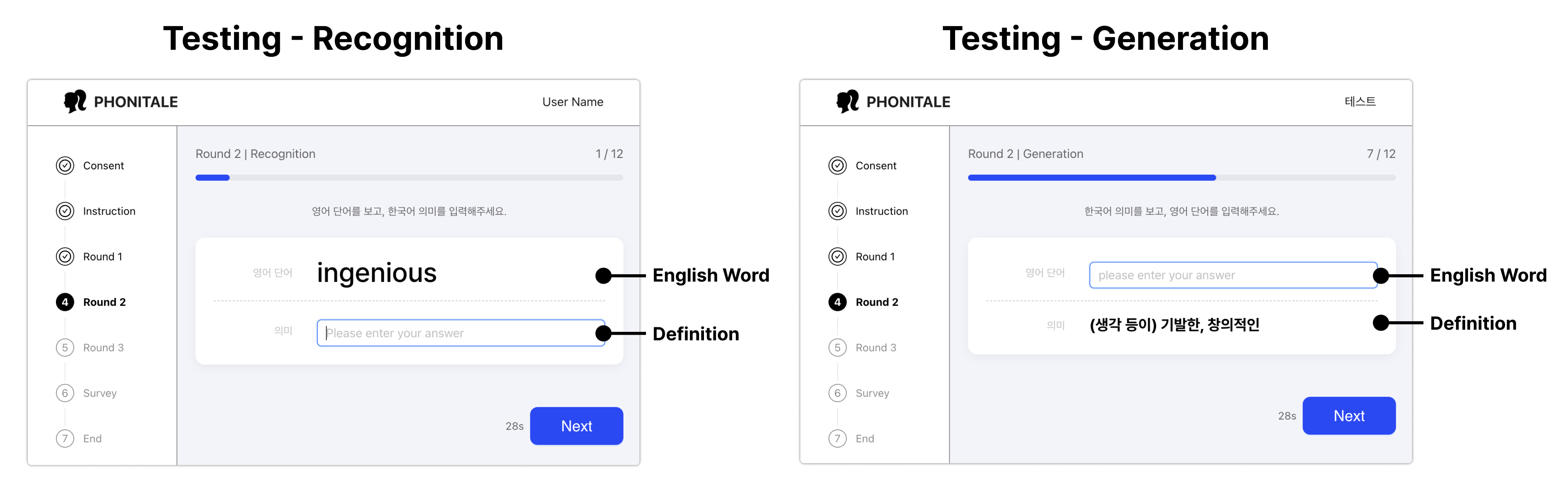}
    \captionof{figure}{User interface for the testing phase. Left: recognition task, Right: generation task.}
    \label{fig:ui-testing}
\end{minipage}

\begin{minipage}{\textwidth}
    \centering
    \includegraphics[width=1.0\textwidth]{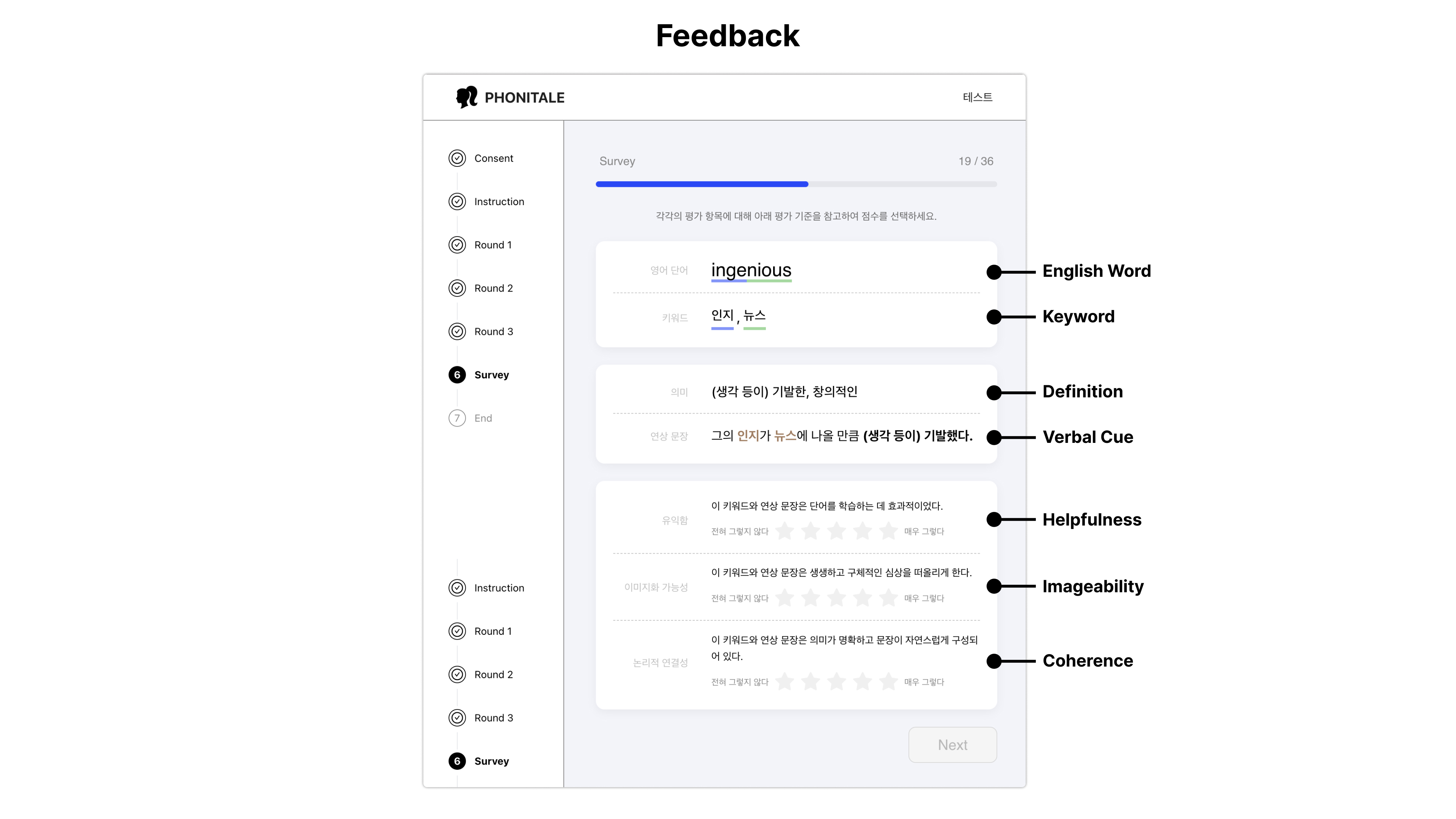}
    \captionof{figure}{User interface for the feedback phase. \\ Participants rate each cue on helpfulness, imageability, and coherence using a 5-point scale.}
    \label{fig:ui-feedback}
\end{minipage}

\subsubsection{Criteria}
\label{appendix:criteria}

Guidelines for rating the 5-point Likert scale used to evaluate \textit{helpfulness}, \textit{coherence}, and \textit{imageability} of verbal cues.
\vspace{1em}
\FloatBarrier

\begin{minipage}{\textwidth}
\centering
\renewcommand{\arraystretch}{1.2}
\begin{tabular}{>{\raggedright}m{2.2cm} >{\raggedright\arraybackslash}m{10cm}}
\toprule
\textbf{Scale} & \textbf{Explanation} \\
\midrule
High (5) & 구조적으로 잘 연관되어 있고, 반복 학습 없이도 단어의 의미를 쉽게 떠올릴 수 있음 \\
         & \textit{\small Well-structured and strongly linked; supports effortless recall without repeated study} \\
\midrule
Medium (3) & 단서와 의미 사이에 약한 연결 고리가 있으나, 기억에 오래 남기엔 부족함 \\
           & \textit{\small Somewhat related, but not strongly memorable or sufficient for long-term recall} \\
\midrule
Low (1) & 단어의 뜻과 단서 사이에 직접적 연결이 거의 없어 기억하거나 학습하는 데 실질적인 도움이 되지 않음 \\
        & \textit{\small Weak or no clear link between cue and word meaning; offers little learning support} \\
\bottomrule
\end{tabular}
\captionof{table}{Instructions for rating \textbf{helpfulness} of the verbal cues.}
\label{tab:scoring_helpfulness}
\end{minipage}

\vspace{2em}
\FloatBarrier

\begin{minipage}{\textwidth}
\centering
\renewcommand{\arraystretch}{1.2}
\begin{tabular}{>{\raggedright}m{2.2cm} >{\raggedright\arraybackslash}m{10cm}}
\toprule
\textbf{Scale} & \textbf{Explanation} \\
\midrule
High (5) & 논리, 어휘, 의미 흐름이 매끄럽고 자연스럽게 구성됨 \\
         & \textit{\small Logically coherent and lexically natural; overall meaning flows well} \\
\midrule
Medium (3) & 비교적 자연스럽지만, 문법이나 논리 흐름에서 약간 부자연스러움 \\
           & \textit{\small Fairly natural but with some grammatical or logical awkwardness} \\
\midrule
Low (1) & 문장이 어색하고 단어 해석과 연결성이 부족함 \\
        & \textit{\small Grammatically awkward or incoherent; lacks clarity or connection to the word} \\
\bottomrule
\end{tabular}
\captionof{table}{Instructions for rating \textbf{coherence} of the verbal cues.}
\label{tab:scoring_coherence}
\end{minipage}

\vspace{2em}
\FloatBarrier

\begin{minipage}{\textwidth}
\centering
\renewcommand{\arraystretch}{1.2}
\begin{tabular}{>{\raggedright}m{2.2cm} >{\raggedright\arraybackslash}m{10cm}}
\toprule
\textbf{Scale} & \textbf{Explanation} \\
\midrule
High (5) & 익숙한 이미지로 쉽게 시각화되며 장면이 구체적으로 떠오름 \\
         & \textit{\small Evokes vivid and specific scenes using familiar imagery} \\
\midrule
Medium (3) & 단어와 관련된 이미지가 조금 있으나 모호하거나 약함 \\
           & \textit{\small Partially imageable, but vague or weakly related to the word} \\
\midrule
Low (1) & 장면이나 상황이 전혀 그려지지 않음 \\
        & \textit{\small Difficult or impossible to visualize any meaningful scene or context} \\
\bottomrule
\end{tabular}
\captionof{table}{Instructions for rating \textbf{imageability} of the verbal cues.}
\label{tab:scoring_imageability}
\end{minipage}
\FloatBarrier

\subsection{Evaluation Materials by Group}

\subsubsection{English Words Set}
\renewcommand{\arraystretch}{1.3} 
\begin{longtable}{|l|l|}
\hline
\textbf{English Word} & \textbf{Korean Definition} \\
\hline
albeit & 비록 $\sim$이기는 하나 \\
\hline
annihilate & 전멸시키다, 붕괴하다 \\
\hline
canny & 약삭빠른, 영리한 \\
\hline
esoteric & 비법을 이어받은, 소수만 이해하는 \\
\hline
incumbent & 의무로서 지워지는, 현직의, 재임 중인 \\
\hline
insolvent & 파산한 \\
\hline
meddlesome & 참견하기를 좋아하는 \\
\hline
probation & 집행유예, 보호관찰, 수습 \\
\hline
reckon & ($\sim$라고) 생각하다, 예상하다 \\
\hline
refurbish & (방, 건물 등을) 새로 꾸미다, 쇄신하다 \\
\hline
resuscitate & 소생시키다 \\
\hline
upheaval & 대변동, 격변 \\
\hline
anachronism & 시대착오, 시대착오적인 사람(관습생각) \\
\hline
bureaucracy & 관료정치, 관료체제, 관료 \\
\hline
delirium & 정신 착란, 헛소리 \\
\hline
demolish & 때려 부수다, 파괴하다, 철거하다 \\
\hline
felon & 중죄인, 흉악범 \\
\hline
incarcerate & 투옥하다, 감금하다 \\
\hline
ingenious & (생각 등이) 기발한, 창의적인 \\
\hline
recidivist & 상습범 \\
\hline
redoubtable & 가공할, 무시무시한, 경외할 만한 \\
\hline
remunerate & 보상하다, 보수를 주다 \\
\hline
render & (어떤 상태가 되게) 만들다, $\sim$하게 하다, 주다, 제출하다 \\
\hline
repercussion & 영향, 파급효과 \\
\hline
autopsy & (사체의) 부검 \\
\hline
congenital & 타고난, 선천적인 \\
\hline
fictitious & 허구의, 지어낸 \\
\hline
inebriate & 취하게 하다, 술꾼 \\
\hline
insurrection & 폭동, 반란 \\
\hline
intransigent & 비협조적인, 비타협적인 \\
\hline
inveterate & (습관 등이) 뿌리 깊은, 고질적인 \\
\hline
mayhem & 대혼란, 아수라장 \\
\hline
peccable & 과오를 범하기 쉬운 \\
\hline
provisional & 임시의, 일시적인 \\
\hline
reimburse & 갚다, 상환하다 \\
\hline
squander & 낭비하다 \\
\hline
\caption{English words used in the experiment with their corresponding Korean definitions.}
\label{tab:appendix-korean-defs}
\end{longtable}

\subsubsection{KSS Keyword and Verbal Cue}
The 36 English words used for human-authored mnemonics were selected from 경선식 영단어 공편토~\cite{kyungsun2020}. These selections are used solely for research and evaluation purposes.
\begin{longtable}{|l|p{0.15\linewidth}|p{0.45\linewidth}|}
\hline
\textbf{English Word} & \textbf{KSS Keyword} & \textbf{KSS Verbal Cue} \\
\hline
albeit & 모두, B (all, B)& 비록 학점이 모두 B이기는 하나 다음에는 A를 받을 것이다. (Although all grades are B, one will get an A next time.) \\
\hline
annihilate & 언, 아이, 끄집어내다 (frozen, child, rescue) & 얼음 속에 언 아이를 밖으로 끄집어내려고 얼음을 붕괴하다 (To rescue a frozen child outside, the ice must be annihilated.) \\
\hline
canny & 캐니 (dig up) & 선거에서 상대방 약점을 캐니? 즉, 약삭빠른, 영리한 (Do you "dig up" the opponent's weakness? That is, shrewd and clever.)\\
\hline
esoteric & 애써, 때리 (to strive, to hit) & 요리의 대가가 제자에게 "애 써!" 하고 때리며 자식에게만 가르쳐주는 비법 (A master chef says "Work hard!" while hitting the apprentice, teaching secret skills only to his own child.) \\
\hline
incumbent & 수입, 번 (income, earned) & 수입을 버는 것은 가장의 의무이기에 짤리지 말고 현직에 있어야 한다 (Earning income is a duty of the head of household, so one must remain in office.) \\
\hline
insolvent & 안, 쌀, 타다 (in, rice, burnt) & 곳간 안에 쌀이 다 타버려 파산한 (In the granary, all the rice has burnt, leaving one insolvent.)\\
\hline
meddlesome & 중간, 몇 (middle, some) & 사람들 사이 중간에 몇 번씩 나타나 참견하기 좋아하는 (Someone who repeatedly appears in the middle of people, eager to meddle.) \\
\hline
probation & 풀어, 뵈이션 (to release, to see) & 풀어주고 또 범행을 하는지 뵈이도록 지켜보는 것. 즉 집행유예, 보호관찰 (Release someone and see whether they commit another crime. That is, probation, parole.) \\
\hline
reckon & 내 껀 (mine) & 이 금도끼는 내 껀 아니라고 생각하다 (Thinking "this golden axe is not mine.") \\
\hline
refurbish & 다시, 퍼, 빛이 (again, to scoop, light is) & 다시 물을 퍼서 빛이 나게 씻어 새로 꾸미다 (Scoop water again to wash and make it shine — to refurbish.) \\
\hline
resuscitate & 다시, 서시다 (again, stand up) & 죽은 사람이 다시 일어서시게 하다. 즉 소생시키다 (To make a dead person rise again — to resuscitate.) \\
\hline
upheaval & 위, 엉덩이 (up, hip) & 거인이 엉덩이를 위로 들고 방귀를 세게 뀌자 사방이 진동하며 일어나는 대변동 (When a giant lifts up his hip and farts loudly, causing upheaval with vibrations everywhere.) \\
\hline
anachronism & 아내, 끌어(내다)니즘 (wife, to drag out-ism) & 칠거지악을 범했다며 아내를 집에서 끌어내는 시대착오적인 생각 (The anachronistic idea of dragging one's wife out of the house, claiming she committed the seven grounds for divorce.) \\
\hline
bureaucracy & 행정부, 크라시 (govern, meaningless word) (bureau, govern) & 의회나 정당이 아닌 행정부로(행정부의 관료들이) 통치하는 관료정치 (Bureaucracy where the executive branch (bureaucrats) govern instead of the parliament or political parties.) \\
\hline
delirium & 닐리리아 (nil-li-ri-a) & "닐리리아" 하며 미쳐서 노래 부르는 정신 착란 (Mental delirium of singing "nil-li-ri-a" while going crazy.) \\
\hline
demolish & 뒤, 말리시다 (behind, stop) & 술에 취해 물건들을 때려 부수는 사람을 뒤에서 말리시다 (Stop someone from behind who is drunkenly demolishing things.) \\
\hline
felon & 팰, 놈 (to beat, guy) & 곤장을 팰 놈, 즉 중죄인, 흉악범 (A guy to be beaten with a rod, that is, a felon, a criminal.) \\
\hline
incarcerate & 안, 칼, 쓰래이! (in, cangue, put on!) & "감옥 안에서 칼을 써!" 하고 투옥하다, 감금하다 ("Put on the cangue in prison!" while incarcerating, confining.) \\
\hline
ingenious & 안, 지니다 (in, to possess) & 머리 안에 지녔어, 기발한 창의적인 생각을 (Possessing ingenious, creative thoughts in one's head.) \\
\hline
recidivist & 다시, 씨디, 비슷 (again, CD, similar) & 불법복제로 처벌받은 후 re(다시) CD를 비슷하게 불법복제하는 상습범 (A recidivist who, after being punished for illegal copying, again makes similar illegal CD copies.) \\
\hline
redoubtable & 다시, 다, 울어 (again, all, cry) & 영화에서 무시무시한 귀신이 나와 아이들이 다시 다 울어 (In the movie, a redoubtable ghost appears and makes all the children cry again.) \\
\hline
remunerate & 뒤, 물어내다 (again, to pay back) & 어떤 대가로 되돌려 물어내다. 즉 보상하다 (To pay back in return for something — to remunerate.) \\
\hline
render & 낸다 (to pay) & 심부름센터에 돈을 낸다. 그리고 ~하게 하다 (Pay money to an errand center. And to render/make something.) \\
\hline
repercussion & 뒤, 퍼지다, 쿠션 (back, spread, cushion) & 공이 뒤로 튀며 쿠션 효과를 내는 파급효과 (The repercussion effect of a ball bouncing back with a cushion effect.) \\
\hline
autopsy & 오!, 톱, 보다 (oh!, saw, see) & 오! 톱으로 시체를 잘라 자세히 보는 부검 (Oh! An autopsy where a corpse is cut with a saw to see in detail.) \\
\hline
congenital & 큰, 제니, 털 (big, Jenny, hair) & 제니라는 사람의 얼굴에 난 큰 털은 타고난, 선천적인 (The big hair on Jenny's face is congenital, inborn.) \\
\hline
fictitious & 픽!, 튀셨수 (pick!, ran away) & 허구의 보물선 사업으로 돈을 끌어모은 뒤 픽! 튀셨수 (After gathering money with a fictitious treasure ship business, pick! he ran away.) \\
\hline
inebriate & 안, 이불이, 에잇! (in, blanket, damn!) & 너무 취해서 술집 안에서 "이불이 어딨지? 에잇! 그냥 바닥에서 자자" 하는 술꾼 (A drunkard so inebriated that in the bar he says "Where's the blanket? Damn! Let's just sleep on the floor.") \\
\hline
insurrection & 안에, 서, 액션 (in, stand, action) & 안에 누워 있는 자들이여, 일어서 액션을 취합시다!하며 폭동을 일으키다 (Those lying inside, let's stand up and take action! starting an insurrection.) \\
\hline
intransigent & 안, 넘어오다, 전투 (in, cross into, battle) & 국경 안으로 넘어와 전투할 만큼 비타협적인 (Intransigent enough to cross into the border and battle.) \\
\hline
inveterate & 안, 뱉어 (in, spit) & 입 안에 침을 자꾸 뱉는 습관이 뿌리 깊은 (Having a deep-rooted habit of constantly spitting saliva from the mouth.) \\
\hline
mayhem & 매인, 햄 (tied to, ham) & 줄에 매인 햄을 서로 먹으려는 대혼란 (Mayhem of everyone trying to eat the ham tied to a string.) \\
\hline
peccable & 팩, 꺼, 불 (pack, turn off, fire) & 한석봉의 어머니가 팩! 하고 불을 꺼 한석봉이 붓글씨를 쓸 때 과오를 범하기 쉬운 (Han Seokbong's mother goes "pack!" and turns off the fire, making Han Seokbong peccable when writing calligraphy.) \\
\hline
provisional & 프로그램, 비잖아 (program, it is empty) & 방송 사고로 지금 내보낼 TV 프로가 비잖아, 임시의 방송이라도 내보내! (Due to a broadcast accident, there's no TV program to air now, so broadcast something provisional!) \\
\hline
reimburse & 다시, 안, 버스 (again, into, bus) & 버스비를 안 내고 내려서 다시 버스 안으로 들어가 버스비를 갚다, 상환하다 (After getting off without paying the bus fare, going back into the bus to reimburse the bus fare.) \\
\hline
squander & 습관, 더 (habit, more) & 돈을 필요한 양보다 더 쓰는 습관, 즉 낭비하다 (The habit of spending more money than needed — to squander.) \\
\hline
\caption{KSS Keyword and Verbal Cue}
\label{tab:KSS keyword and verbal cue}
\end{longtable}

\subsubsection{OGR Keyword and Verbal Cue}
\begin{longtable}{|l|p{0.15\linewidth}|p{0.45\linewidth}|}
\hline
\textbf{English Word} & \textbf{OGR Keyword} & \textbf{OGR Verbal Cue} \\
\hline
albeit & 얼, 비 (freeze, rain) & 비록 $\sim$이기는 하나 그는 얼어붙은 길을 걸으며 비를 맞았다. (Although ~, he walked on the frozen road in the rain.) \\
\hline
annihilate & 안, 아이, 라이트 (inside, child, light) & 안에서 아이들이 놀던 건물이 붕괴하자 라이트가 깜빡였다. (When the building where children were playing inside collapsed, the light flickered.) \\
\hline
canny & 케냐 (Kenya) & 케냐에서 영리한 사업 확장. (Clever business expansion in Kenya.) \\
\hline
esoteric & 에서, 테이크 (from, take) & 식당에서 비법을 이어받은 요리를 테이크아웃했다. (Took out food that inherited secret recipes from the restaurant.) \\
\hline
incumbent & 인, 금, 반지 (person, gold, ring) & 재임 중인 그는 인과 금 반지를 받았다. (The incumbent received a person and gold ring.) \\
\hline
insolvent & 인, 솔, 벤트 (person, Sol, bent) & 인생에서 솔직함을 추구하다 벤처 사업으로 파산한. (Bankrupt from venture business while pursuing honesty in life.) \\
\hline
meddlesome & 메달, 섬 (medal, island) & 참견하기를 좋아하는 그녀는 메달 수여식에 섬까지 갔다. (She who likes to meddle went to the island for the medal ceremony.) \\
\hline
probation & 포, 배, 신 (four, ship, god) & 포트폴리오를 배포할 수습 기자가 신문사에서 일했다. (A probationary reporter who would distribute portfolios worked at the newspaper company.) \\
\hline
reckon & 레고, 큰 (LEGO, big) & 레고로 큰 성을 만들 수 있다고 생각했다. (Thought he could build a big castle with LEGO.) \\
\hline
refurbish & 리, 버스 (Li, bus) & 리 작업실을 쇄신하여 버스 정류장에서 보는 예술 공간으로 만들었다. (Renovated Li's workshop into an art space visible from the bus stop.) \\
\hline
resuscitate & 리, 서, 시, 테이프 (Li, stand, hour, tape) & 리 박사는 환자를 소생시키려 서둘러 시계를 보며 테이프를 확인했다. (Dr. Li hurriedly checked the tape while looking at the clock to resuscitate the patient.) \\
\hline
upheaval & 업, 이불 (industry, blanket) & 경제적 업 속에서도 이불 속 불안은 대변동의 징조였다. (Even amid economic industry, the anxiety under the blanket was a sign of great upheaval.) \\
\hline
anachronism & 아, 낙지, 룬, 이슴 (ah, octopus, rune, -ism) & 아, 낙지를 룬으로 변신시키려다 시대착오로 끝났다. (Ah, trying to transform octopus into runes ended in anachronism.) \\
\hline
bureaucracy & 비, 오, 러, 크시 (rain, come, Rue, -ksi) & 비가 오면 러시아워처럼 되는 관료정치. (Bureaucracy that becomes like rush hour when it rains.) \\
\hline
delirium & 딜러, 이음 (dealer, joint) & 딜러가 이음 없이 헛소리를 했다. (The dealer spoke deliriously without pause.) \\
\hline
demolish & 디, 머리, 시 (D, head, city) & 건물을 파괴하여 디테일을 무시하고 머리 속 시각을 그린다. (Demolish buildings, ignore details, and draw the vision in one's head.) \\
\hline
felon & 펠, 언니 (Pel, sister) & 펠과 그의 언니와 함께한 흉악범의 계획. (The felon's plan with Pel and his sister.) \\
\hline
incarcerate & 인형, 칼, 새 (doll, knife, bird) & 인형과 칼을 가진 새를 지키려다 투옥되고 만다. (Ended up imprisoned trying to protect the bird with a doll and knife.) \\
\hline
ingenious & 인기, 뉴스 (popularity, news) & 인기를 끌며 뉴스에 나온 기발한 아이디어. (Ingenious idea that gained popularity and appeared on the news.) \\
\hline
recidivist & 리, 시, 디, 피스트 (Li, hour, D, fist) & 리 마을의 상습범이 시계탑 근처에서 디자이너 가방을 훔치고 피스트를 벌였다. (The recidivist from Li village stole a designer bag near the clock tower and had a fist fight.) \\
\hline
redoubtable & 래, 다운, 더블 (Rae, down, double) & 무시무시한 래퍼는 다운된 무대에서 더블 타임 랩을 했다. (The formidable rapper performed double-time rap on the downed stage.) \\
\hline
remunerate & 리, 무, 내다 (Li, nothing, give out) & 리 씨는 무더위 속의 노고를 보상했다. (Mr. Li compensated for the hard work in the sweltering heat.) \\
\hline
render & 랜, 돌 (LAN, stone) & 랜을 돌로 던져 마법을 만들다. (Throw the LAN with stone to create magic.) \\
\hline
repercussion & 리본, 커피, 션 (ribbon, coffee, Shawn) & 리본을 달고 커피를 마신 션의 하루에 긍정적인 영향. (Positive impact on Shawn's day wearing a ribbon and drinking coffee.) \\
\hline
autopsy & 옷, 합시다 (clothes, let's do) & 사체의 옷을 확인하고 부검을 합시다. (Let's check the corpse's clothes and perform an autopsy.) \\
\hline
congenital & 컨, 제니, 탈 (Con, Jenny, mask) & 컨과 제니는 탈을 쓰고 타고난 무대를 빛냈다. (Con and Jenny wore masks and shone on their natural stage.) \\
\hline
fictitious & 피크, 티셔츠 (peak, T-shirt) & 피크닉 티셔츠에 허구의 이야기를 담았다. (Put fictitious stories on the picnic T-shirt.) \\
\hline
inebriate & 인, 애비, 에이 (person, father, A) & 술꾼 인영은 애비와 에이급 바에서 시간을 보냈다. (Drunkard Inyoung spent time with father at an A-grade bar.) \\
\hline
insurrection & 인, 수레, 션 (person, cart, Shawn) & 폭동 중 인파 속에서 수레를 밀고 션이 이끌었다. (During the insurrection, Shawn led by pushing a cart through the crowd.) \\
\hline
intransigent & 인, 트랜스, 젠트 (person, trance, gent) & 인적이 드문 길에서 비협조적인 그는, 트랜스 음악에 젠트하게 반응했다. (The intransigent person on the deserted road reacted gently to trance music.) \\
\hline
inveterate & 인, 배터리 (person, battery) & 인내심이 뿌리 깊은 철수는 배터리가 소모될 때까지 참았다. (Cheolsu, with deep-rooted patience, endured until the battery was depleted.) \\
\hline
mayhem & 메기, 힘 (catfish, strength) & 메기가 힘을 쓰자 대혼란이 일어났다. (When the catfish used its strength, mayhem broke out.) \\
\hline
peccable & 배, 꺼, 불 (ship, turn off, fire) & 과오를 범하기 쉬운 그는 배를 타고 가다가 전등을 꺼서 불이 꺼졌다. (He who was prone to error turned off the light while on the ship, and the fire went out.) \\
\hline
provisional & 프로, 비서, 널 (pro, secretary, you) & 프로 프로젝트에서 비서로 임시의 역할을 했다. (Played a provisional role as secretary in the pro project.) \\
\hline
reimburse & 림, 버스 (rim, bus) & 림에서 버스를 기다리며 돈을 갚다. (Repay money while waiting for the bus at the rim.) \\
\hline
squander & 숯, 권, 더 (charcoal, volume, more) & 숯을 권 단위로 더 사서 낭비했다. (Wasted by buying more charcoal by the volume.) \\
\hline
\caption{OGR Keyword and Verbal Cue}
\label{tab:OGR keyword and verbal cue}
\end{longtable}

\subsubsection{PHT Keyword and Verbal Cue}
\begin{longtable}{|l|p{0.15\linewidth}|p{0.45\linewidth}|}
\hline
\textbf{English Word} & \textbf{PHT Keyword} & \textbf{PHT Verbal Cue} \\
\hline
albeit & 올, 바이트 (all, byte) & 비록 올바른 바이트 이기는 하나 그는 망설였다. (Although it was the correct byte, he hesitated.) \\
\hline
annihilate & 얼결, 레이더 (accidentally, radar) & 얼결에 레이더 시스템이 붕괴했다. (The radar system accidentally collapsed.) \\
\hline
canny & 케어, 니은 (care, nieun) & 그는 케어를 받으며 니은을 그릴 때도 영리했다. (He was clever even when drawing 'nieun' while receiving care.) \\
\hline
esoteric & 애, 스프링 (child, spring) & 그 애는 스프링의 작동 원리를 소수만 이해하는 전문가였다. (That child was an expert in spring operation principles understood by only a few.) \\
\hline
incumbent & 인어, 콘센트 (mermaid, outlet) & 인어가 콘센트 옆에서 재임 중인 느낌으로 노래했다. (The mermaid sang with a feeling of being incumbent next to the outlet.) \\
\hline
insolvent & 인어, 선발대 (mermaid, advance team) & 인어가 선발대를 따라가다 파산했다. (The mermaid went bankrupt while following the advance team.) \\
\hline
meddlesome & 매달다, 섬 (hang, island) & 그는 섬을 매달고 싶다며 모든 일에 참견하기를 좋아했다. (He liked to meddle in everything, saying he wanted to hang the island.) \\
\hline
probation & 프로, 봉변 (pro, trouble) & 그는 프로처럼 봉변을 수습했다. (He handled the trouble like a pro.) \\
\hline
reckon & 레게, 컨셉 (reggae, concept) & 그는 레게 스타일이 멋진 컨셉이라고 생각했다. (He thought reggae style was a cool concept.) \\
\hline
refurbish & 리포터, 쉬 (reporter, rest) & 리포터는 쉬지 않고 방을 쇄신했다. (The reporter renovated the room without rest.) \\
\hline
resuscitate & 리더, 스케이트 (leader, skate) & 리더는 스케이트를 통해 팀의 사기를 소생시켰다. (The leader revived the team's morale through skating.) \\
\hline
upheaval & 업, 바이블 (industry, bible) & 회사의 업계 바이블이 격변을 맞았다. (The company's industry bible faced upheaval.) \\
\hline
anachronism & 어느, 크리스천 (any, Christian) & 어느 시대에나 크리스천을 시대착오적인 사람(관습생각)이라 부르는 경우가 있다. (In any era, there are cases where Christians are called anachronistic people.) \\
\hline
bureaucracy & 병따개, 러시아 (bottle opener, Russia) & 병따개를 들고 러시아에 간 관료. (A bureaucrat who went to Russia with a bottle opener.) \\
\hline
delirium & 달리아, 라임 (dahlia, lime) & 달리아 꽃밭에서 라임을 곁들이며 헛소리를 늘어놓았다. (He rambled deliriously in the dahlia garden with lime.) \\
\hline
demolish & 대물림, 쉬 (inheritance, rest) & 대물림된 집을 쉬지 않고 철거했다. (He demolished the inherited house without rest.) \\
\hline
felon & 펠레우스, 런던 (Peleus, London) & 펠레우스를 런던으로 데려간 흉악범. (The felon who took Peleus to London.) \\
\hline
incarcerate & 인가, 샐러드 (approval, salad) & 그는 인가를 받지 못해 샐러드를 훔치다 투옥되었다. (He was imprisoned for stealing salad because he couldn't get approval.) \\
\hline
ingenious & 인지, 뉴스 (cognition, news) & 그의 인지가 뉴스에 나올 만큼 (생각 등이) 기발했다. (His cognition was ingenious enough to make the news.) \\
\hline
recidivist & 리시버, 비슷이 (receiver, similarly) & 리시버를 통해 비슷이 행동하는 상습범이 있었다. (There was a recidivist who acted similarly through the receiver.) \\
\hline
redoubtable & 라디오, 토플 (radio, TOEFL) & 라디오에서 들려오는 무시무시한 뉴스가 토플 준비에 방해가 되었다. (The formidable news from the radio interfered with TOEFL preparation.) \\
\hline
remunerate & 라면, 레이더 (ramen, radar) & 그는 라면을 먹고 레이더를 고치면 보상하겠다고 말했다. (He said he would compensate if they ate ramen and fixed the radar.) \\
\hline
render & 랜드, 더 (land, more) & 그는 랜드를 더 재미있게 만들었다. (He made the land more interesting.) \\
\hline
repercussion & 리그, 파티션 (league, partition) & 리그가 파티션에 미친 영향은 절대적이었다. (The league's impact on the partition was absolute.) \\
\hline
autopsy & 오, 텃세 (oh, territorial behavior) & 오랜만에 텃세를 부리는 이웃이 (사체의) 부검 이야기로 사람들을 놀라게 했다. (The neighbor showing territorial behavior after a long time surprised people with autopsy stories.) \\
\hline
congenital & 컨테이너, 털 (container, fur) & 그는 컨테이너를 열며 털을 정리하는 솜씨가 타고났다. (He had a natural talent for organizing fur while opening containers.) \\
\hline
fictitious & 픽, 투표소 (pick, polling station) & 픽을 던진 후 투표소에서 모든 이야기를 지어냈다. (After throwing the pick, he made up all the stories at the polling station.) \\
\hline
inebriate & 일없이, 리야드 (idly, Riyadh) & 그는 일없이 리야드에서 모든 사람을 취하게 했다. (He idly made everyone drunk in Riyadh.) \\
\hline
insurrection & 인어, 선입견 (mermaid, prejudice) & 인어는 선입견에 맞서 반란을 일으켰다. (The mermaid rebelled against prejudice.) \\
\hline
intransigent & 인터넷, 지진대 (internet, seismic zone) & 그는 인터넷에서 지진대 정보를 찾으면서도 비타협적인 태도를 유지했다. (He maintained an intransigent attitude while searching for seismic zone information on the internet.) \\
\hline
inveterate & 인어, 배터리 (mermaid, battery) & 인어의 배터리 사용 습관은 뿌리 깊었다. (The mermaid's battery usage habit was deeply rooted.) \\
\hline
mayhem & 매입, 힘 (purchase, strength) & 매입으로 인해 힘이 생기면서 아수라장이 되었다. (The purchase gave strength and caused mayhem.) \\
\hline
peccable & 패, 커플 (faction, couple) & 그는 새로운 패를 내놓고 커플 앞에서 과오를 범하기 쉬운 사람임을 드러냈다. (He revealed himself to be someone prone to error in front of the couple while presenting a new faction.) \\
\hline
provisional & 패러디, 저널 (parody, journal) & 패러디 저널은 일시적인 인기를 끌었다. (The parody journal gained temporary popularity.) \\
\hline
reimburse & 레임덕, 스무 (lame duck, twenty) & 레임덕 시기에도 그는 스무 번이나 빚을 갚았다. (Even during the lame duck period, he repaid debts twenty times.) \\
\hline
squander & 세관, 더 (customs, more) & 그는 세관에서 시간이 더 낭비되었다. (He wasted more time at customs.) \\
\hline
\caption{PHT Keyword and Verbal Cue}
\label{tab:PHT keyword and verbal cue}
\end{longtable}

\subsection{Metrics}
\subsubsection{LLM-as-a-judge Prompt}
\label{appendix:LLM-as-a-judge Prompt}

The prompt was originally designed in Korean. For reproducibility, we provide both the original and its English translation.

\begin{longtable}{|p{2cm}|p{11cm}|}
\hline
Prompt & 당신은 영어 어휘 학습을 평가하는 채점자입니다. \\
& \small\textit{You are a grader evaluating English vocabulary learning.} \\
 & 정답 의미: 게으른, 나태한 \\
 & \small\textit{Correct meaning: lazy, idle} \\
 & 학습자 응답: 게으른 \\
 & \small\textit{Learner response: lazy} \\
 & 학습자의 응답이 정답 의미와 일치하는지 평가해주세요. \\
 & \small\textit{Please evaluate whether the learner's response matches the correct meaning.} \\
 & 다른 표현이나 다른 품사로 설명했더라도 의미가 유사하다면 정답으로 인정합니다. \\
 & \small\textit{Accept as correct if the meaning is similar, even if expressed differently or in a different part of speech.} \\
 & 1(정답) 또는 0(오답)으로만 응답하세요. \\
 & \small\textit{Respond only with 1 (correct) or 0 (incorrect).} \\
\hline
Response & 1 \\
\hline
\caption{Prompts for evaluating correctness of recognition responses.}
\label{tab:eval_recognition_prompt}
\end{longtable}

\begin{longtable}{|p{2cm}|p{11cm}|}
\hline
Prompt & 당신은 영어 어휘 학습을 평가하는 채점자입니다. \\
& \small\textit{You are a grader evaluating English vocabulary learning.} \\
 & 정답 영단어: squander \\
 & \small\textit{Correct English word: squander} \\
 & 학습자 응답: squander \\
 & \small\textit{Learner response: squander} \\
 & 학습자의 응답이 정답과 일치하는지 평가해주세요. 약간의 오타, 대소문자 차이, 복수형/단수형 차이, 품사 차이 등은 허용합니다. \\
 & \small\textit{Please evaluate whether the learner's response matches the correct answer. Minor typos, case differences, plural/singular differences, and part of speech differences are allowed.} \\
 & 하지만 같은 뜻을 가지는 다른 영단어는 오답으로 판단합니다. \\
 & \small\textit{However, different English words with the same meaning are considered incorrect.} \\
 & 1(정답) 또는 0(오답)으로만 응답하세요. \\
 & \small\textit{Respond only with 1 (correct) or 0 (incorrect).} \\
\hline
Response & 1 \\
\hline
\caption{Prompts for evaluating correctness of generation responses.}
\label{tab:eval_generation_prompt}
\end{longtable}

\subsection{Case Study}
\label{appendix: Case Study}

\subsubsection{Correctness Comparison by Word}

\FloatBarrier
\begin{minipage}{\textwidth}
\centering
\includegraphics[width=1.0\textwidth]{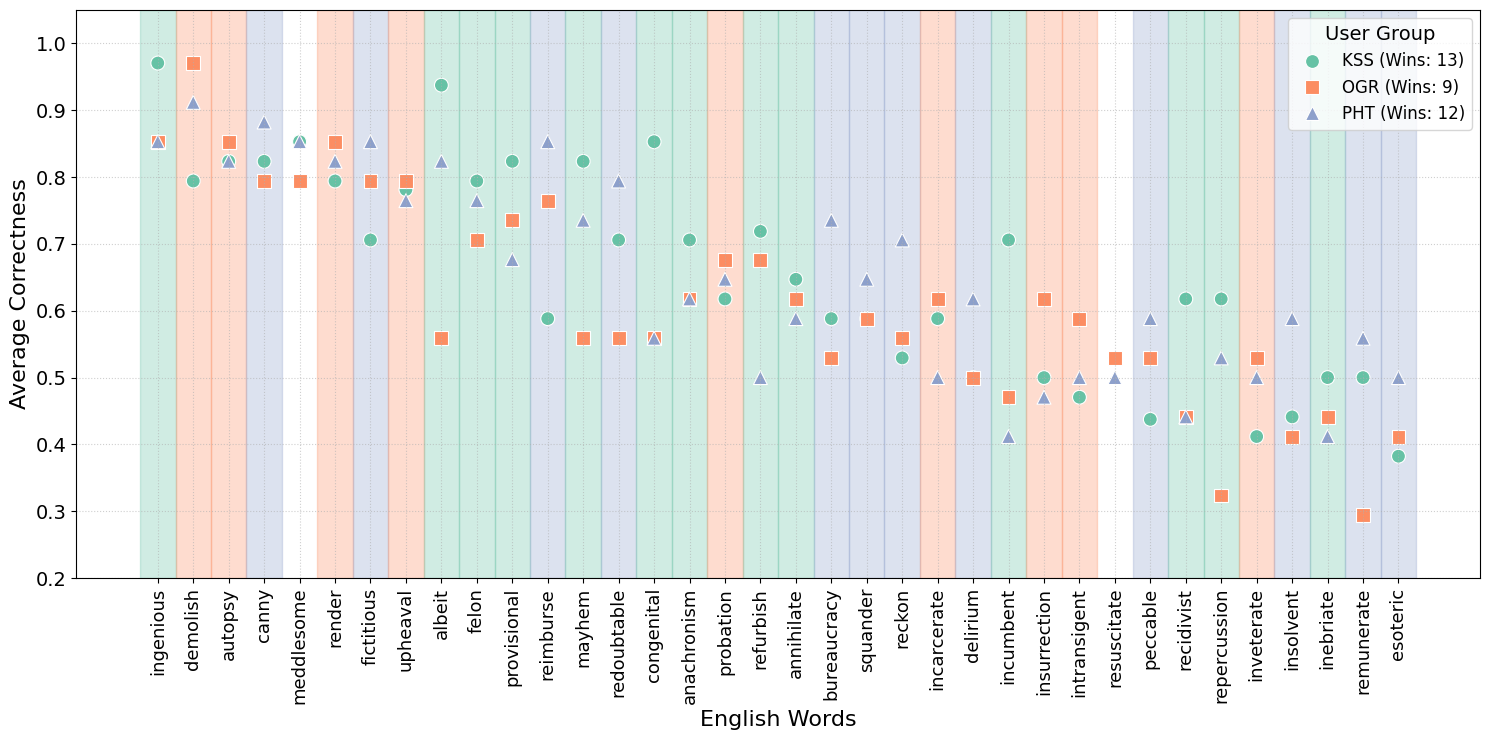}
\captionof{figure}{Per-word correctness scores across user groups. Each marker represents the average correctness score for a specific target word, grouped by user condition (KSS, OGR, PHT). For each word, the highest-scoring group is highlighted with a shaded background. The number of wins is comparable between \textsc{KSS} (13) and \textsc{PHT} (12), suggesting no consistent dominance across all words. Given the variation in group performance by word, this pattern motivates further analysis into the word-level characteristics that influence correctness across different groups.}
\label{fig:appendix-correctness-wordlevel}
\end{minipage}

\subsubsection{Qualitative Comparison of PHT and KSS}

\FloatBarrier
\begin{minipage}{\textwidth}
\centering
\setlength{\tabcolsep}{12pt} 
\renewcommand{\arraystretch}{1.2}
\begin{tabularx}{\textwidth}{
    >{\hsize=0.5\hsize\RaggedRight}X  
    >{\hsize=0.8\hsize\RaggedRight}X  
    >{\hsize=1.2\hsize\RaggedRight}X  
    >{\hsize=0.8\hsize\RaggedRight}X  
    >{\hsize=1.5\hsize\RaggedRight\arraybackslash}X 
}
\toprule
\textbf{Word \newline (IPA)} & \textbf{PHT Keyword \newline (IPA)} & \textbf{PHT Verbal Cue} & \textbf{KSS Keyword \newline (IPA)} & \textbf{KSS Verbal Cue} \\
\midrule
\textit{reckon} \newline \textipa{/"rEk@n/}
& \textit{레게, 컨셉} \newline \textipa{/l E g E/}, \newline \textipa{/k\textsuperscript{h} 2 n s E p/}
& 그는 레게 스타일이 멋진 컨셉이라고 생각했다. \newline {\small\itshape He thought reggae style was a cool concept.}
& \textit{내 껀} \newline \textipa{/n E k\textsuperscript{h} 2 n/}
& 이 금도끼는 내 껀(내 것) 아니라고 생각했다. \newline {\small\itshape He thought this golden axe was not mine.} \\
\midrule
\textit{render} \newline \textipa{/"rEnd@r/}
& \textit{랜드, 더} \newline \textipa{/l E n d 1/}, \newline \textipa{/t 2/}
& 그는 랜드를 더 재미있게 만들었다. \newline {\small\itshape He made the land more fun.}
& \textit{낸다} \newline \textipa{/n E n d a/}
& 심부름센터에 돈을 낸다(주다, 제출하다). 그리고 ~하게 하다. \newline {\small\itshape He pays money to the errand center (to give, to submit), and thus makes something happen.} \\
\bottomrule
\end{tabularx}
\captionof{table}{Examples where \textsc{PHT} outperformed \textsc{KSS}, based on correctness rankings. IPA shown beneath keyword sequence.}
\label{tab:appendix-pht-kss-strong}
\end{minipage}

\FloatBarrier
\begin{center}
\begin{minipage}{\textwidth}
\setlength{\tabcolsep}{12pt} 
\renewcommand{\arraystretch}{1.2}
\begin{tabularx}{\textwidth}{
    >{\hsize=0.5\hsize\RaggedRight}X  
    >{\hsize=0.8\hsize\RaggedRight}X  
    >{\hsize=1.2\hsize\RaggedRight}X  
    >{\hsize=0.8\hsize\RaggedRight}X  
    >{\hsize=1.5\hsize\RaggedRight\arraybackslash}X 
}
\toprule
\textbf{Word} & \textbf{PHT Keyword} \newline \textit{· POS \newline · Similarity} & \textbf{PHT Verbal Cue} & \textbf{KSS Keyword} \newline \textit{· POS \newline · Similarity} & \textbf{KSS Verbal Cue} \\
\midrule
\textit{felon} 
& \textit{펠레우스, 런던} \newline · Noun, Noun \newline · 0.86
& 펠레우스를 런던으로 데려간 흉악범. \newline {\small\itshape The vicious criminal who took Peleus to London.}
& \textit{팰, 놈} \newline · Verb, Noun \newline · 1.15
& 곤장을 팰 놈, 즉 중죄인, 흉악범. \newline {\small\itshape The one to be beaten with a cudgel—that is, a serious criminal or felon.} \\
\midrule
\textit{mayhem} 
& \textit{매입, 힘} \newline · Noun, Noun \newline · 1.07
& 매입으로 인해 힘이 생기면서 아수라장이 되었다. \newline {\small\itshape Because of the purchase, strength arose and chaos ensued.}
& \textit{매인, 햄} \newline · Adj, Noun \newline · 1.10
& 줄에 매인 햄을 서로 먹으려고 수많은 개들이 대혼란, 아수라장. \newline {\small\itshape Countless dogs fought to eat the ham tied to a rope, causing great turmoil and mayhem.} \\
\bottomrule
\end{tabularx}
\captionof{table}{Examples where \textsc{KSS} outperformed \textsc{PHT}, based on correctness rankings. Part-of-speech and phonetic similarity annotated.}
\label{tab:appendix-pht-kss-weak}
\end{minipage}
\end{center}

\clearpage
\appendix

\end{CJK}
\end{document}